\documentclass[a4paper,12pt,english]{article}

\usepackage[utf8]{inputenc} 
\usepackage[T1]{fontenc}    
\usepackage[x11names]{xcolor}
\usepackage{graphicx}
\usepackage{amsmath}
\usepackage{amssymb}
\usepackage{amsthm}
\usepackage{dsfont}
\usepackage{algorithm}
\usepackage{algpseudocode}
\usepackage{geometry}
\usepackage{colortbl}
\usepackage[authoryear,round]{natbib}
\usepackage{tikz}
\usetikzlibrary{tikzmark, shapes.arrows}
\usepackage{caption}
\usepackage{subcaption}
\usepackage{booktabs}

\newcommand{\R}{\mathbb{R}}
\newcommand{\E}{\mathbb{E}}
\newcommand{\x}{\mathbf{x}}

\newcommand{\z}{\mathbf{z}}
\newcommand{\w}{\mathbf{w}}
\newcommand{\uu}{\mathbf{u}}

\newcommand{\bs}[1]{\boldsymbol{\mathbf{#1}}}
\newcommand{\monthyeardate}{\ifcase \month \or January\or February\or March\or %
April\or May \or June\or July\or August\or September\or October\or November\or %
December\fi, \number \year}

\newtheorem{definition}{Definition}

\newtheorem{remark}{Remark}

\graphicspath{{}}

\title{Distributional encoding for Gaussian process regression with qualitative inputs}

\author{%
  Sébastien Da Veiga \\
  ENSAI, CREST - UMR 9194, Rennes, France\\
  \texttt{sebastien.da-veiga@ensai.fr} \\  
}

\date{}

\begin{document}

\maketitle

\begin{abstract}
Gaussian Process (GP) regression is a popular and sample-efficient approach for many engineering applications, where observations are expensive to acquire, and is also a central ingredient of Bayesian optimization (BO), a highly prevailing method for the optimization of black-box functions. However, when all or some input variables are categorical, building a predictive and computationally efficient GP remains challenging. Starting from the naive target encoding idea, where the original categorical values are replaced with the mean of the target variable for that category, we propose a generalization based on \textit{distributional encoding} (DE) which makes use of all samples of the target variable for a category. To handle this type of encoding inside the GP, we build upon recent results on characteristic kernels for probability distributions, based on the maximum mean discrepancy and the Wasserstein distance. We also discuss several extensions for classification, multi-task learning and incorporation or auxiliary information. Our approach is validated empirically, and we demonstrate state-of-the-art predictive performance on a variety of synthetic and real-world datasets. DE is naturally complementary to recent advances in BO over discrete and mixed-spaces.
\end{abstract}

\section{Introduction}

Gaussian Process (GP) regression is a powerful nonparametric Bayesian approach for modeling complex functions, particularly in settings where data are scarce or expensive to obtain. Its ability to provide uncertainty estimates alongside predictions makes it especially attractive for computer experiments, physical simulations, and other scientific applications where sample efficiency is critical. In these contexts, datasets are often small, and the underlying relationships can be highly nonlinear. GP regression has thus become a standard tool in fields such as engineering design, climate modeling and materials science, where it is used to build surrogate models or emulators of computationally intensive simulators \citep{marrel2008efficient,rohmer2022,zhang2020}. 

\medskip

Despite the flexibility of GP models for continuous inputs, incorporating qualitative or categorical variables poses a significant challenge due to the lack of a natural metric space structure for such variables. Standard GP formulations rely on kernel functions that assume inputs lie in a continuous space where notions of distance or similarity are well-defined. However, categorical inputs, such as material types, experimental settings or system configurations, consist of unordered and discrete levels that cannot be straightforwardly embedded into such spaces. Several classes of approaches have been developed to overcome this limitation. 
A first strategy involves \emph{dedicated kernels} for categorical variables, such as the Hamming kernel \citep{song2012feature} or ANOVA decomposition kernels \citep{storlie2011surface}, which define similarity based on level matching. These methods are easy to implement and computationally efficient, and they retain the full structure of GP inference. However, they often assume that each level is equally dissimilar from the others, ignoring domain-specific similarities, and they often lack adaptability to complex structures among levels.
A more flexible alternative is \emph{covariance parameterization}, where the correlation or covariance matrix across levels of the categorical variable is treated as a parameter to be estimated. This can be achieved, for example, by assigning a symmetric positive semi-definite matrix of free parameters to the categorical dimension, enabling the model to learn similarities between levels from data. While this approach can lead to significantly improved performance by capturing complex dependencies, it introduces a potentially large number of additional parameters (growing quadratically with the number of levels). Structured parameterizations, such as low-rank decompositions, are often needed to make these models identifiable and stable. 
A third strategy involves \emph{latent variable models}, where each level of a categorical variable is associated with a continuous latent representation, typically learned jointly with the GP model \citep{zhang2020,oune2021latent}. These models enable smooth interpolation, are naturally compatible with continuous input kernels, and can uncover interpretable structures among levels. However, the latent space is often non-identifiable, and inference can be computationally intensive. Finally, standard machine learning \emph{encodings} such as one-hot encoding or target encoding allow categorical variables to be plugged into off-the-shelf kernels by embedding them into a Euclidean space. One-hot encoding can lead to high-dimensional sparse representations, which are manageable for small numbers of categories but scale poorly and usually yields a model with low accuracy.

\medskip

Building upon target encoding, we introduce the idea of using \emph{distributional encoding} to represent qualitative inputs. The key idea is to associate each level of a categorical variable with a probability distribution over the output or some auxiliary data space, and to define similarity between levels via a kernel between these distributions. This enables the incorporation of side information associated with each level, such as observed responses from lower fidelity models, or experimental data. Advances in kernel methods on distributions leverage tools such as maximum mean discrepancy (MMD) kernels, which embed distributions into a reproducing kernel Hilbert space (RKHS) and define similarity via inner products in that space. These kernels are computationally efficient and benefit from strong theoretical guarantees. Another powerful approach uses Wasserstein distances, derived from optimal transport theory, to define kernels that account for the geometry of probability distributions. These methods can be used to define similarities between empirical distributions associated with categorical levels, providing a principled and expressive way to incorporate qualitative information. Nevertheless, practical implementation requires careful design choices, including computational approximations, and ensuring that the resulting kernels remain positive definite.

\medskip

This paper is structured as follows. In Section 1, we provide a brief overview of GP regression and review standard approaches for handling qualitative inputs. Section 2 introduces the concept of kernels on probability distributions and their application to qualitative variables, discussing several families of such kernels. We also introduce various extensions, which widen the scope of our proposal. Section 3 illustrates the performance of these methods on datasets representative of engineering applications, comparing them to existing approaches and highlighting the benefits and limitations of each.

\section{Gaussian process regression with qualitative inputs}

We focus on a standard regression setting with an input vector $\w=(\x,\uu)$ of dimension $p+q$ with $p$ quantitative inputs $\x\in\R^{p}$ and $q$ qualitative inputs $\uu$, and an output $y=f(\w)+\epsilon\in\R$ where $f$ is the unknown regression function and $\epsilon \sim \mathcal{N}(0,\eta^2)$ is a Gaussian random noise. We further assume that each qualitative input $u_t$ has levels denoted $1,\ldots,L_t$ for $t=1,\ldots,q$. From an i.i.d. training sample $(\w^{(i)},y^{(i)})_{i=1}^{n}$ of size $n$, our goal is to infer the unknown regression function $f$.

\medskip

In GP regression, a Gaussian process prior is placed on function $f$:
\[f(w) \sim GP(m(\w),k(\w,\w’))\]
with mean function $m$ and positive semi-definite kernel $k$. It follows that the joint distribution of the observed target values and the function values $\mathbf{f_*}:=(f(\w_*^{(i)}))_{i=1}^{N_*}$ at new test locations $\w_*=(\w_*^{(i)})_{i=1}^{N_*}$ writes (see, \textit{e.g.}, \citet{gpml})
\begin{equation}
    \begin{bmatrix}
    \mathbf{y}\\
    \mathbf{f}_*
    \end{bmatrix}
    \sim \mathcal{N} 
    \left( 
    \begin{bmatrix}
    m(\w) \\
    m(\w_*)
    \end{bmatrix}, 
    \begin{bmatrix}
    \mathbf{K} + \eta^2 \mathbf{I} & \mathbf{K}_*^T\\
    \mathbf{K}_* & \mathbf{K}_{**}
    \end{bmatrix}
    \right)\,,
\end{equation}
where $\w=(\w^{(i)})_{i=1}^{n}$, $\mathbf{y}=(y^{(i)})_{i=1}^{n}$ and  $\mathbf{K}$, $\mathbf{K}_{**}$, $\mathbf{K}_{*}$ are the train, test and test/train Gram matrices from kernel $k$, respectively. The posterior distribution of $\mathbf{f}_*$, obtained by conditioning the joint distribution on the observed data, is also Gaussian: $\mathbf{f}_* | \w, \mathbf{y}, \w_* \sim \mathcal{N}(\mathbf{\bar{m}}(\w_*),\mathbf{\bar{\Sigma}})$ with mean and variance given by
\begin{align*}
\mathbf{\bar{m}}(\w_*) &= m(\w_*) + \mathbf{K}_*(\mathbf{K}+\eta^2\mathbf{I})^{-1} \mathbf{y}\,, \\
\mathbf{\bar{\Sigma}}&=\mathbf{K}_{**}-\mathbf{K}_* (\mathbf{K}+\eta^2 \mathbf{I})^{-1}\mathbf{K}_*^T\,.
\end{align*}
The mean of this posterior distribution is used as an estimate of the regression function $f$. When all features are quantitative, it is common to use a tensorized product kernel
\[k(\w,\w') = k_{\textrm{cont}}(x_1,x'_1) \times \ldots \times k_{\textrm{cont}}(x_p,x'_p)\]
with $k_{\textrm{cont}}$ any univariate kernel such as the Gaussian or the Mat\'{e}rn one. For a qualitative input with levels $l=1,\ldots,L$, defining a positive semi-definite kernel is equivalent to defining a positive semi-definite (psd) matrix $\mathbf{T}\in\R^{L\times L}$ with entries
\[[\mathbf{T}]_{ij} = k_{\textrm{cat}}(i,j)\]
for $i,\, j=1,\ldots,L$ and where $k_{\textrm{cat}}$ is a categorical kernel measuring the similarity of levels $i$ and $j$. A tensorized product kernel with both quantitative and qualitative features is then given by
\begin{align}
k(\w,\w') = k_{\textrm{cont}}(x_1,x'_1) \times \ldots \times k_{\textrm{cont}}(x_p,x'_p) \times [\mathbf{T}_1]_{u_1,u’_1}\times\ldots\times [\mathbf{T}_q]_{u_q,u’_q}. \label{eq:tensor}
\end{align}
In practice, the kernel is parameterized by lengthscales, that must be estimated with training data either by maximum likelihood or more recent robust alternatives based on priors \citep{gu2018robust}.

\medskip

The most direct way to accommodate qualitative inputs is to build a kernel that operates exclusively on category levels. The simplest instance is the dirac (or Hamming) kernel \(k_{\textrm{cat}}(u_t,u'_t)=\mathds{1}_{u_t=u'_t}\), which can be directly plugged in (\ref{eq:tensor}) \citep{song2012feature}. A similar kernel was also proposed by \citet{gower1971general}. Besides the tensorized product form, ANOVA‐style kernels split the covariance into main categorical effects, pure continuous effects, and their interactions, see, e.g.,  \citet{storlie2011surface}. In practice, however, the most popular approaches rely on sparse reparameterization of the psd matrices $\mathbf{T}_j$ or latent variables, which we detail in the following sections.

\subsection{Covariance parameterization}

Covariance parameterization methods assign each categorical variable its own psd correlation matrix \(\mathbf{T}_j\in\R^{L_j\times L_j} \), with learnable entries. When data are plentiful, maximum-likelihood estimation of \(\mathbf{T}_j\) can uncover nuanced similarities, e.g. two engine prototypes that behave identically at low torque but diverge at high torque, thereby improving predictive accuracy and uncertainty quantification.

\medskip

Because the number of free parameters grows quadratically with the number of levels, more refined strategies are required in practice. Cholesky factorizations \(\mathbf{T}_j=\mathbf{L}\mathbf{L}^T\) with diagonal or spherical reparameterizations enforce positive definiteness while allowing gradient-based optimization \citep{rapisarda2007}. Low-rank or sparse factorizations \(\mathbf{T}=\mathbf{R}\mathbf{R}^T\) with \(\mathbf{R} \in \R^{L_j\times r}\) and rank \(r \ll L_j\) reduce parameter count and often capture the dominant structure with negligible loss, see \citet{kirchhoff2020}.

\medskip

Simpler parameterizations were also proposed, such as compound-symmetry \citep{katz2011}
\[[T]_{ij} = v \delta_{i=j} + c \delta_{i\neq j},\]
which was recently extended with the idea of group-kernels \citep{roustant2020}
\[[T]_{ij} = v_g \delta_{i=j} + c_{g(i),g(j)} \delta_{i\neq j}\]
where $g=1,\ldots,G$ denotes a pre-specified clustering of the levels into groups, $v_g$ is the within-group covariance and $c$ is the between-group covariance between $g(i)$ and $g(j)$, the groups levels $i$ and $j$ belong to. When such group structure is known beforehand from prior knowledge, group-kernels produce impressive predictive performance in practice \citep{rohmer2022}.

\subsection{Latent variables}

Latent-variable techniques embed every categorical level \(l\) into a continuous vector \(\z_l\in\R^d\) and apply any standard kernel \(k_{\textrm{cont}}\) in that latent space. Intuitively, levels that end up close in \(\R^d\) share similar response behaviour: this technique thus introduces a notion of similarity among categorical levels through their proximity in the latent space. The positions of the levels in the latent space are typically learned jointly with the GP hyperparameters via marginal likelihood maximization. Empirical studies show that even two- or three-dimensional embeddings can recover underlying chemical families or material properties, thereby providing domain experts with interpretable "maps" of qualitative design spaces \citep{zhang2020,oune2021latent}. Such latent variable methods are especially appealing when one suspects that the categories lie on or near a continuous manifold, as is often the case in engineering design and material science. Another key advantage of this approach is its scalability and smoothness: latent embeddings reduce the parameter count compared to full covariance matrices and allow for generalization across similar levels. The most prominent method is Latent Variable Gaussian Process (LVGP, \citet{zhang2020}).

\medskip

However, the embeddings are not identifiable up to rotations and scalings, which complicates interpretation, and the likelihood landscape can be multimodal: optimization can stall in local minima or produce embeddings that merely reflect random initialization. This may be a computational hurdle, which we will illustrate in Appendix \ref{sec:addxp}.  In addition, because similarity is now learned indirectly through latent coordinates, extrapolation to unseen levels is impossible without a second-stage model that would predict the new embedding. Intrinsically, these limitations come from the \emph{supervised} nature of these embeddings: a straightforward workaround is to consider instead \emph{unsupervised} or \emph{weakly supervised} embeddings, generally referred to as \emph{encodings}.

\subsection{Standard encoding methods}
\label{sec:base_encod}

Standard encoding methods, such as one-hot encoding and target encoding, are often used as baseline strategies for incorporating qualitative inputs into machine learning models. 

\medskip

One-hot (or dummy) encoding transforms each level of a categorical variable into a binary vector, with each element corresponding to a distinct level. This approach is straightforward and easy to implement, but it imposes a strong assumption of independence between categories. In practice, when used with GPs, one-hot encoding can be implemented in two ways: (a) each binary vector can be considered as a new feature with a specific kernel, but this inflates the problem dimensionality or (b) a continuous kernel on the concatenation of all binary variables can be assembled. In both cases, the resulting kernel will in fact be equivalent to the dirac kernel up to a multiplicative constant. This can severely limit the GP predictive performance, as illustrated in previous work. Note that a recent extension was proposed by \citet{saves2023}, with much better accuracy.

\medskip

Target encoding, on the other hand, replaces each category with a statistic derived from the response variable, typically the mean target value associated with that level: this yields a single continuous feature per categorical variable and preserves the number of GP hyperparameters. More precisely, for each qualitative input $u_t$, $t=1,\ldots,q$, we first denote $N_{tl} = \sum_{i=1}^n \mathds{1}_{u_t^{(i)}=l}$ the number of training samples for which $u_t$ has value $l$ for $l=1,\ldots,L_j$. A sample from $u_t$ with level $l$ is replaced by
\begin{eqnarray*}
\bar{y}_{t,l}=\frac{1}{N_{tl}} \sum_{i=1}^n y^{(i)}\mathds{1}_{u_t^{(i)}=l}
\end{eqnarray*}
for $l=1,\ldots,L_j$, which is an empirical version of the conditional mean $\E(Y\vert u_t=l)$. As an illustration, we will consider an artifical running example: Table \ref{tab:dataset} and Table \ref{tab:tencoding} below show the dataset and the result of vanilla target encoding, where we have replaced $U_1$ by a new virtual continuous input $X_3$ which contains the encodings. After repeating this encoding process for all qualitative inputs, we end up with only continuous inputs (initial ones and virtual ones obtained through encoding) which can be handled by a GP with a Gram matrix defined as
\begin{eqnarray*}
[K]_{ij} = k(\w^{(i)},\w^{(j)}) = \prod_{s=1}^p k_{\textrm{cont}}(x^{(i)}_s,x^{(j)}_s)\prod_{t=1}^q k_{\textrm{cont}}(\bar{y}_{t,u^{(i)}_t},\bar{y}_{t,u^{(j)}_t}).
\end{eqnarray*}

\begin{table*}[hbt!]
\centering
        \large
        \begin{minipage}{0.45\linewidth}
            \centering
            \begin{tabular}{|c|c|c|c|}
\hline
$\mathbf{X_1}$ & $\mathbf{X_2}$ & $\mathbf{U_1}$ & $\mathbf{Y}$ \\
\hline
0.47 & -1.47 & \textcolor{IndianRed3}{red} & -1.5 \\
\hline
0.52 & -0.79 & \textcolor{SeaGreen4}{green} & 0.20 \\
\hline
0.11 & -2.67 & \textcolor{SeaGreen4}{green} & 0.48 \\
\hline
0.75 & 0.43 & \textcolor{SteelBlue4}{blue} & 1.82 \\
\hline
0.11 & 1.91 & \textcolor{IndianRed3}{red} & -4.2 \\
\hline
0.96 & 2.92 & \textcolor{SteelBlue4}{blue} & 2.34 \\
\hline
0.64 & 0.33 & \textcolor{SteelBlue4}{blue} & 4.51 \\
\hline
0.01 & 2.14 & \textcolor{IndianRed3}{red} & -3.7 \\
\hline
0.15 & 1.39 & \textcolor{SeaGreen4}{green} & 0.86 \\
\hline
0.63 & -1.93 & \textcolor{IndianRed3}{red} & -2.9 \\
\hline
\end{tabular} 
\caption{Original dataset.}
            \label{tab:dataset}
        \end{minipage}
        \hspace{1cm}
        \begin{minipage}{.45\linewidth}
            \centering
            \begin{tabular}{|c|c|c|c|}
\hline
$\mathbf{X_1}$ & $\mathbf{X_2}$ & $\mathbf{X_3}$ & $\mathbf{Y}$ \\
\hline
0.47 & -1.47 & \textcolor{IndianRed3}{-3.075} & -1.5 \\
\hline
0.52 & -0.79 & \textcolor{SeaGreen4}{0.51} & 0.20 \\
\hline
0.11 & -2.67 & \textcolor{SeaGreen4}{0.51} & 0.48 \\
\hline
0.75 & 0.43 & \textcolor{SteelBlue4}{2.89} & 1.82 \\
\hline
0.11 & 1.91 & \textcolor{IndianRed3}{-3.075} & -4.2 \\
\hline
0.96 & 2.92 & \textcolor{SteelBlue4}{2.89} & 2.34 \\
\hline
0.64 & 0.33 & \textcolor{SteelBlue4}{2.89} & 4.51 \\
\hline
0.01 & 2.14 & \textcolor{IndianRed3}{-3.075} & -3.7 \\
\hline
0.15 & 1.39 & \textcolor{SeaGreen4}{0.51} & 0.86 \\
\hline
0.63 & -1.93 & \textcolor{IndianRed3}{-3.075} & -2.9 \\
\hline
\end{tabular}
\caption{Target encoding.}
            \label{tab:tencoding}
        \end{minipage}
    \end{table*}

\begin{figure}[hbt!]
\centering
\includegraphics[width=0.8\textwidth]{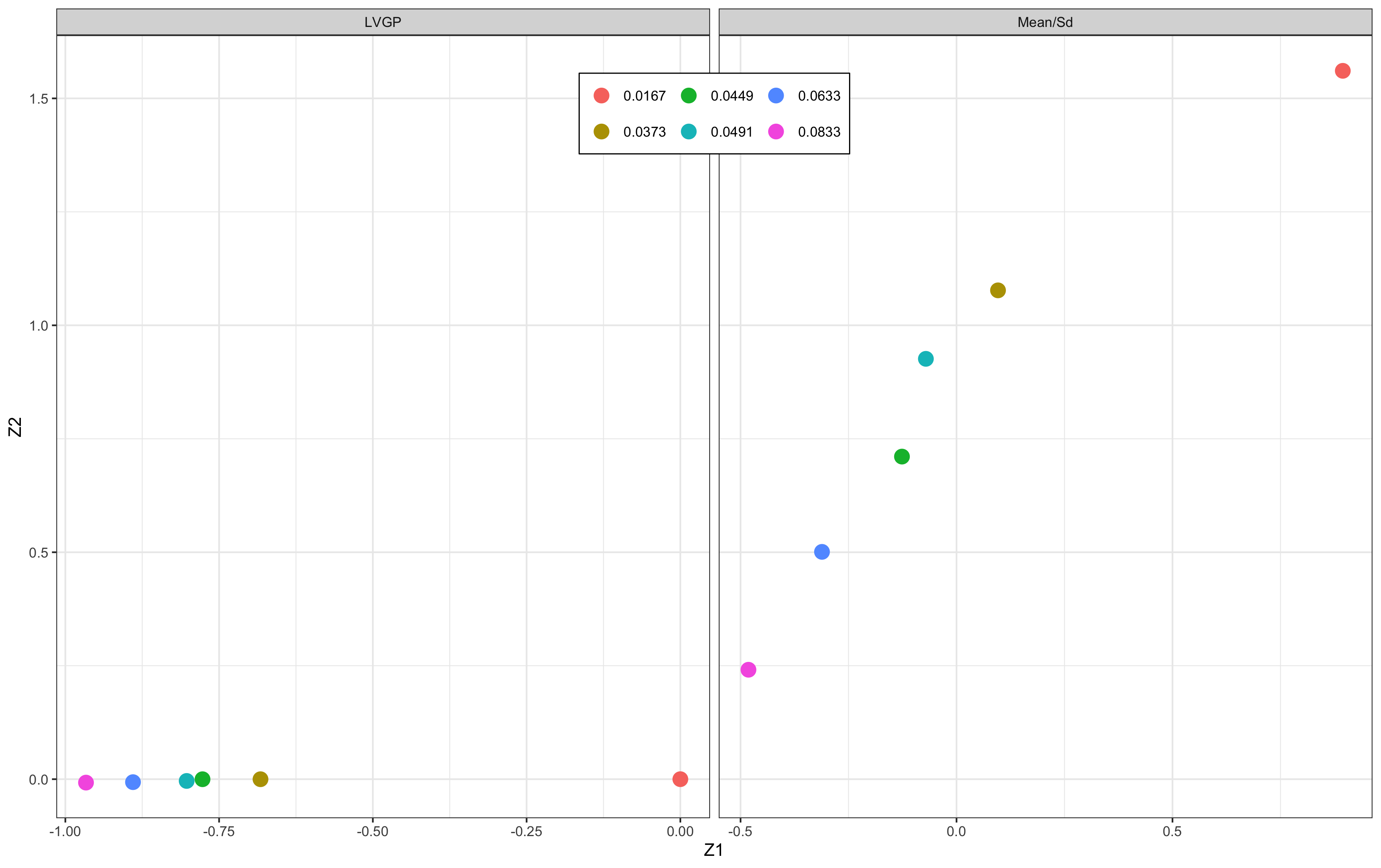} 
\caption{Two-dimensional LVGP embedding versus mean/standard deviation encoding for the beam bending test case.}
\label{fig:beam_encoding}
\end{figure}

A straightforward extension is to consider also additional statistical summaries of the output as new virtual inputs, for instance the standard deviation. For illustration purposes, we compare below a two-dimensional supervised encoding obtained with LVGP and a weakly supervised one consisting of the mean and the standard deviation. Figure \ref{fig:beam_encoding} shows these two encoding for the beam bending case which depends on a unique qualitative input with 6 levels (see Section \ref{sec:xp} for details).

\begin{figure}[hbt!]
\centering        
\includegraphics[width=0.8\textwidth]{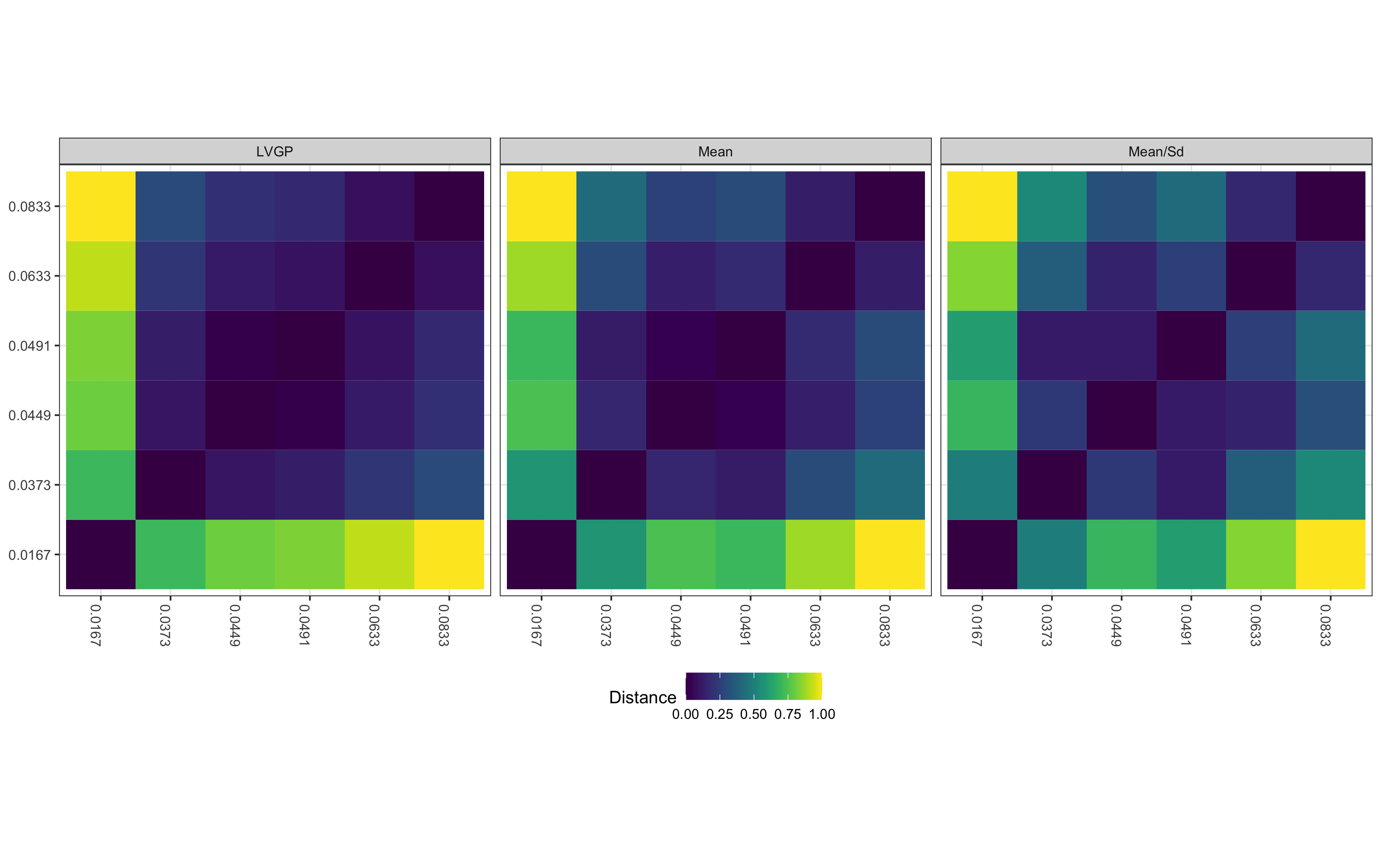} 
\caption{Normalized distance matrix between levels for LVGP, mean and mean/standard deviation encodings for the beam bending test case.}
\label{fig:beam_dist}
\end{figure}

Perhaps surprisingly, we observe that the LVGP embedding and the mean/standard deviation encoding lead to highly similar latent representations, up to a rotation. This is confirmed in Figure \ref{fig:beam_dist}, where we compare the distance matrix between each level representation. Since only this distance matrix is considered within the GP kernel, we expect similar predictive performance for both approaches.

\medskip

To the best of our knowledge, such encodings are rarely evaluated in GP benchmarks for qualitative inputs, although they often yield competitive models, as shown in Section \ref{sec:xp}. However, the mean or the standard deviation are crude summaries of the response, that may fail to capture complex relationships between levels: in the following, we thus propose to account for the entire distribution of the output for each level.

\newpage

\section{Distributional encoding}

A promising extension of target encoding is to represent each qualitative level as a probability distribution over observed responses associated with each level. The key idea is to define a kernel on distributions \( k_{\textrm{P}}(P, Q) \), where each distribution encodes empirical knowledge about a qualitative level. In a similar vein as for mean encoding, we consider
\begin{eqnarray*}
\hat{P}^{Y}_{t,l}=\frac{1}{N_{tl}} \sum_{i=1}^n \delta_{y^{(i)}}\mathds{1}_{u_t^{(i)}=l}
\end{eqnarray*}
where $N_{tl}=\sum_{i=1}^n \mathds{1}_{u_t^{(i)}=l}$ for $l=1,\ldots,L_j$, the empirical version of the conditional probability distribution $P(Y\vert u_t=l)$. Table \ref{tab:dencoding} provides an illustration of this so-called \emph{distributional encoding}.

\begin{table*}[hbt!]
\large
\centering
        \begin{minipage}{0.45\linewidth}
            \centering
            \begin{tabular}{|c|c|c|c|}
\hline
$\mathbf{X_1}$ & $\mathbf{X_2}$ & $\mathbf{U_1}$ & $\mathbf{Y}$ \\
\hline
0.47 & -1.47 & \textcolor{IndianRed3}{red} & -1.5 \\
\hline
0.52 & -0.79 & \textcolor{SeaGreen4}{green} & 0.20 \\
\hline
0.11 & -2.67 & \textcolor{SeaGreen4}{green} & 0.48 \\
\hline
0.75 & 0.43 & \textcolor{SteelBlue4}{blue} & 1.82 \\
\hline
0.11 & 1.91 & \textcolor{IndianRed3}{red} & -4.2 \\
\hline
0.96 & 2.92 & \textcolor{SteelBlue4}{blue} & 2.34 \\
\hline
0.64 & 0.33 & \textcolor{SteelBlue4}{blue} & 4.51 \\
\hline
0.01 & 2.14 & \textcolor{IndianRed3}{red} & -3.7 \\
\hline
0.15 & 1.39 & \textcolor{SeaGreen4}{green} & 0.86 \\
\hline
0.63 & -1.93 & \textcolor{IndianRed3}{red} & -2.9 \\
\hline
\end{tabular}
\caption{Original dataset.}
\label{tab:dataset_regr}
        \end{minipage}
        \hspace{1cm}
        \begin{minipage}{.45\linewidth}
            \centering
            
            \begin{tabular}{|c|c|c|c|}
\hline
$\mathbf{X_1}$ & $\mathbf{X_2}$ & $\mathbf{X_3}$ & $\mathbf{Y}$ \\
\hline
0.47 & -1.47 & \includegraphics[width=1cm, height=0.4cm]{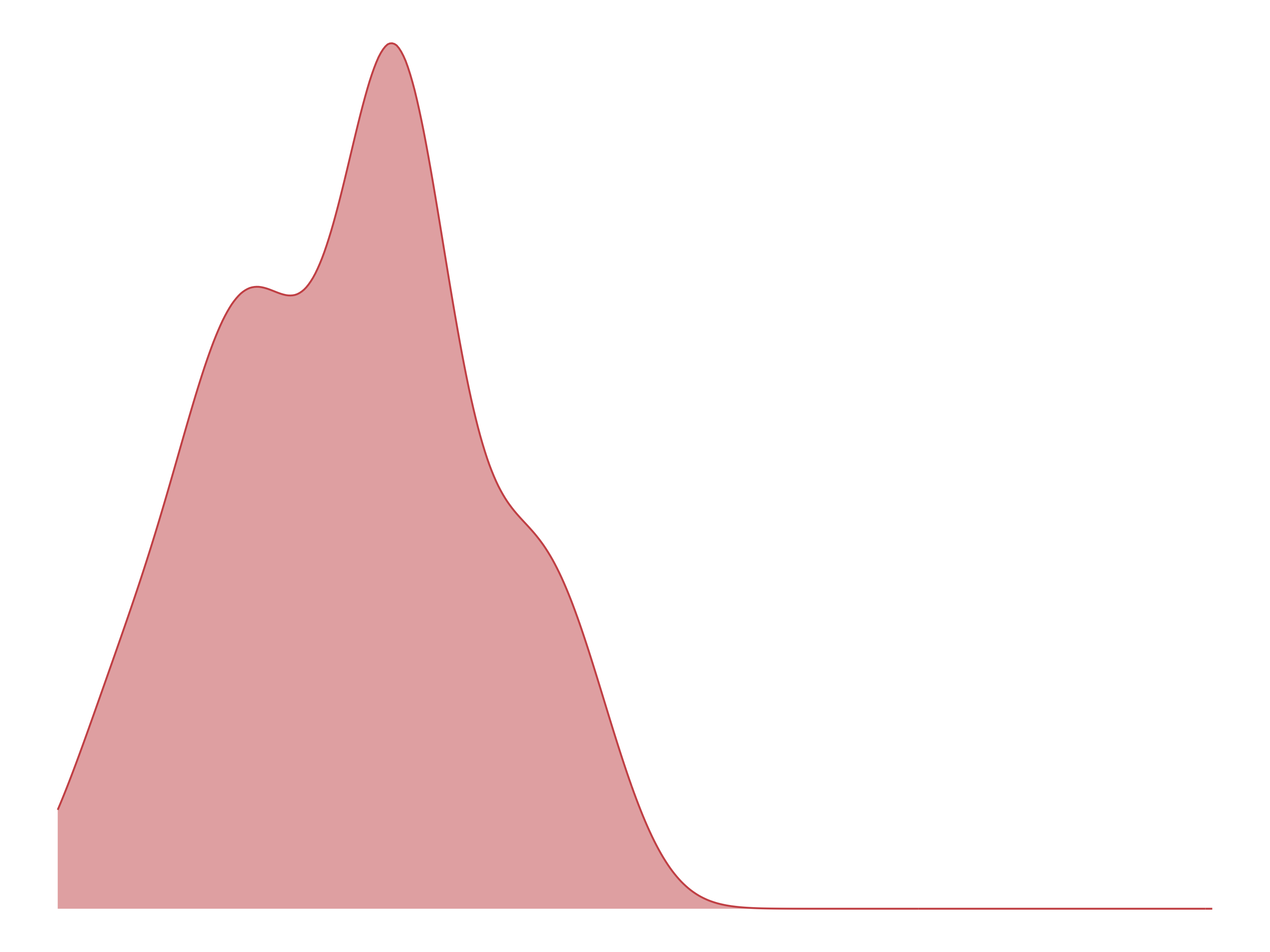} & -1.5 \\
\hline
0.52 & -0.79 & \includegraphics[width=1cm, height=0.4cm]{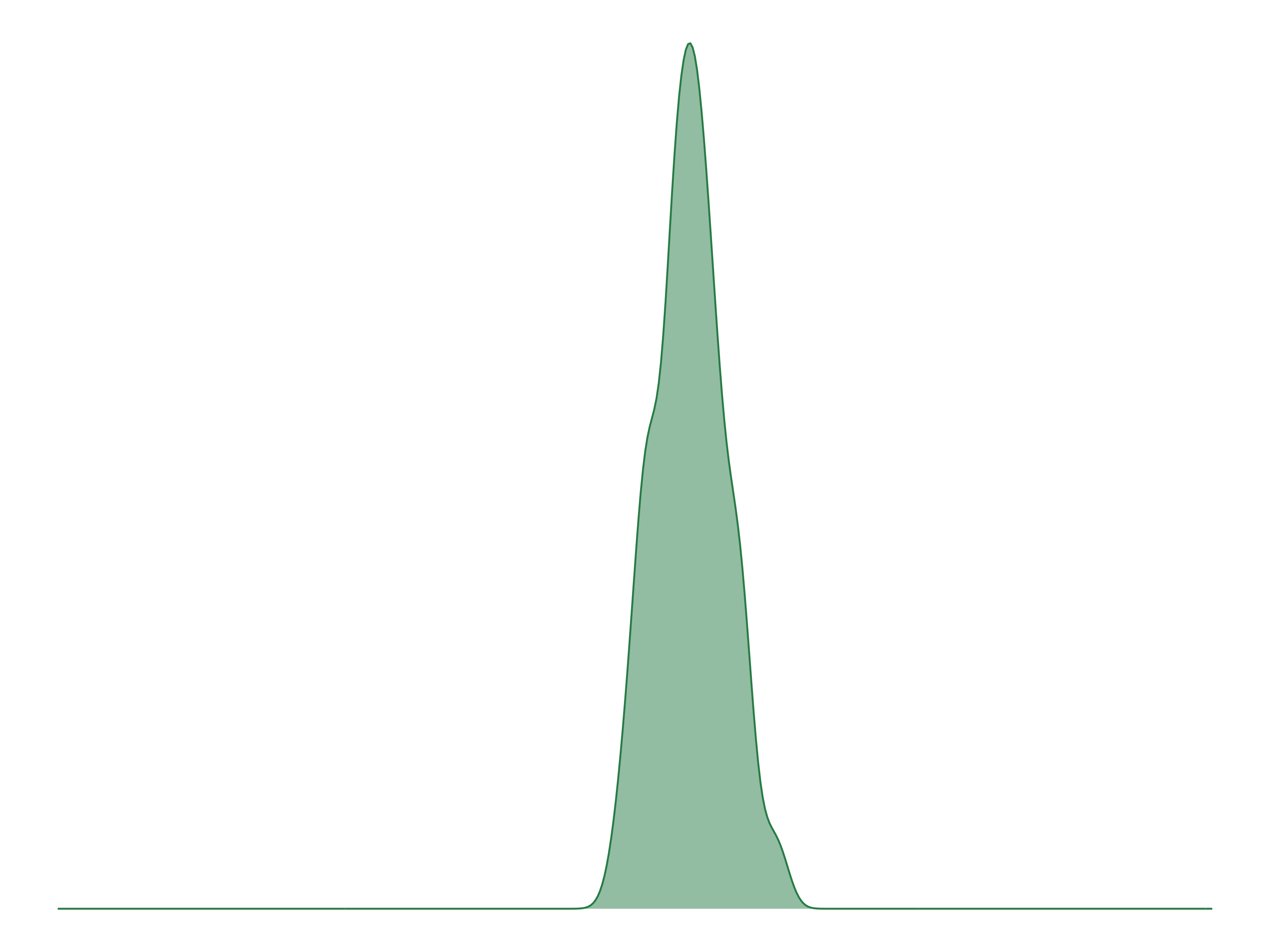} & 0.20 \\
\hline
0.11 & -2.67 & \includegraphics[width=1cm, height=0.4cm]{d2.png} & 0.48 \\
\hline
0.75 & 0.43 & \includegraphics[width=1cm, height=0.4cm]{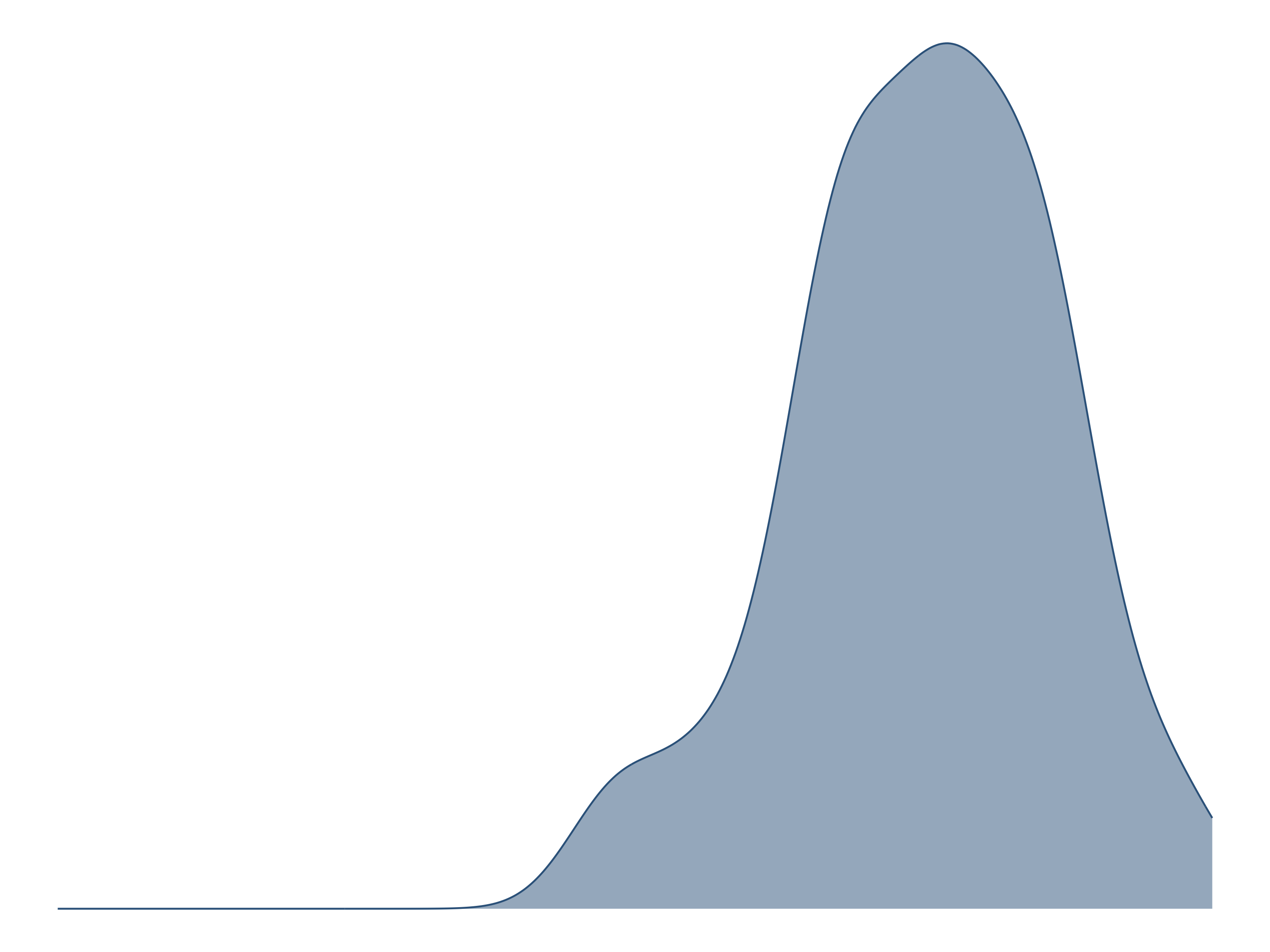} & 1.82 \\
\hline
0.11 & 1.91 & \includegraphics[width=1cm, height=0.4cm]{d1.png} & -4.2 \\
\hline
0.96 & 2.92 & \includegraphics[width=1cm, height=0.4cm]{d3.png} & 2.34 \\
\hline
0.64 & 0.33 & \includegraphics[width=1cm, height=0.4cm]{d3.png} & 4.51 \\
\hline
0.01 & 2.14 & \includegraphics[width=1cm, height=0.4cm]{d1.png} & -3.7 \\
\hline
0.15 & 1.39 & \includegraphics[width=1cm, height=0.4cm]{d2.png} & 0.86 \\
\hline
0.63 & -1.93 & \includegraphics[width=1cm, height=0.4cm]{d1.png} & -2.9 \\
\hline
\end{tabular}
\caption{Distributional encoding.}
            \label{tab:dencoding}
        \end{minipage}
    \end{table*}

\medskip

Equipped with this weakly supervised representation, the last step is to propose a suitable kernel. Focusing on product kernels, the associated Gram matrix we propose is given by
\begin{eqnarray*}
[K]_{ij} = k(\w^{(i)},\w^{(j)}) = \prod_{s=1}^p k_{\textrm{cont}}(x^{(i)}_s,x^{(j)}_s)\prod_{t=1}^q k_{\textrm{P}}(\hat{P}^{Y}_{t,u^{(i)}_t},\hat{P}^{Y}_{t,u^{(j)}_t})
\end{eqnarray*}
where $k_{\textrm{P}}$ is any kernel on probability distributions, but note that any standard kernel combination can be envisioned (e.g., sum, ANOVA, ...). The key ingredient is to define a positive semi-definite kernel on probability distributions which is both computationally efficient and expressive enough.

\subsection{Kernels on probability distributions}\label{sec:kernels}

Several families of kernels defined on probability distributions have emerged in the literature, each with different assumptions and computational properties.

\medskip

\begin{figure}[hbt!]
\centering
\includegraphics[width=0.8\textwidth]{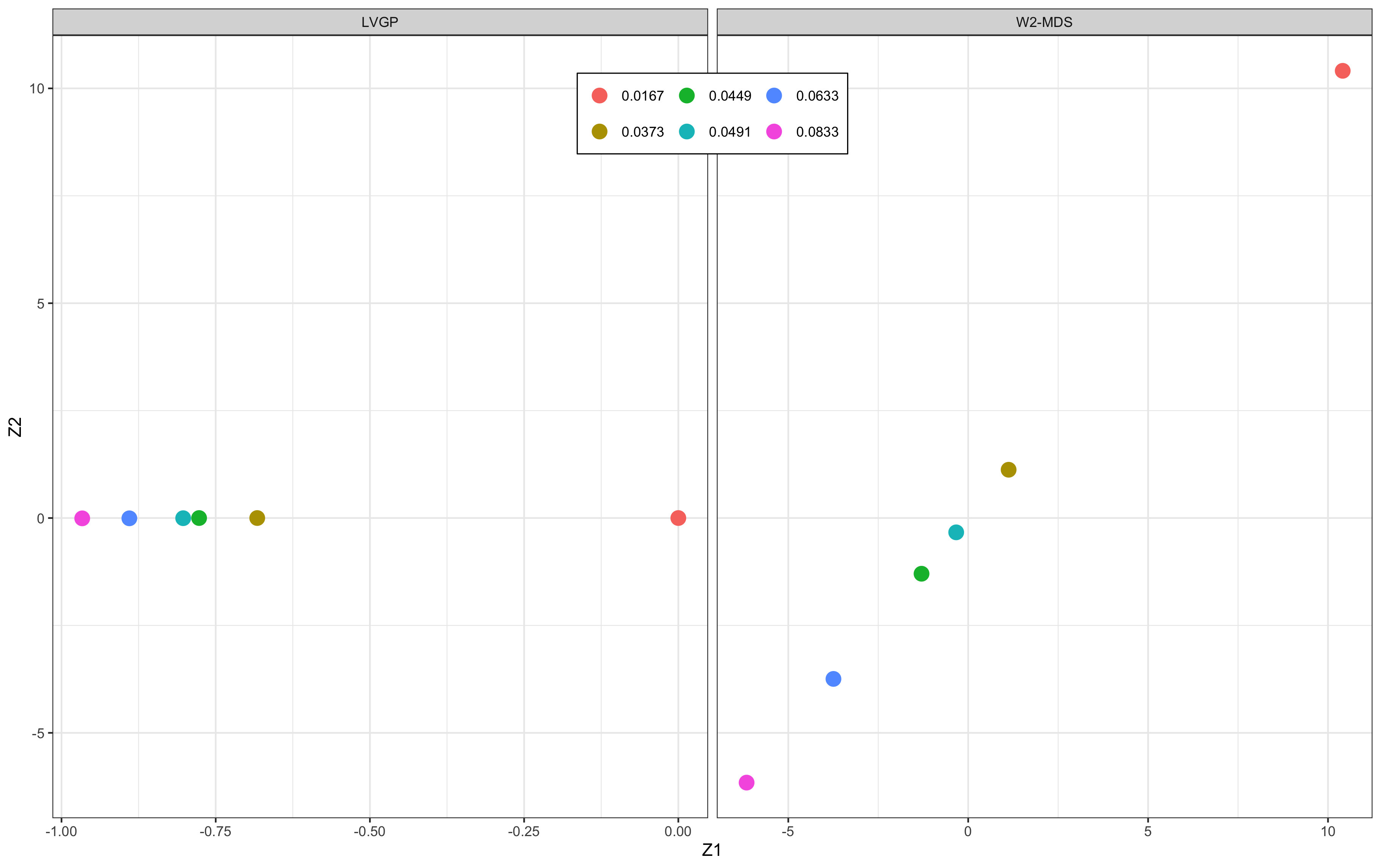} 
\caption{LVGP embedding versus $W_2$ encoding for the beam bending test case.}
\label{fig:beam_encodingW2}
\end{figure}

\begin{figure}[hbt!]
\centering
\includegraphics[width=0.8\textwidth]{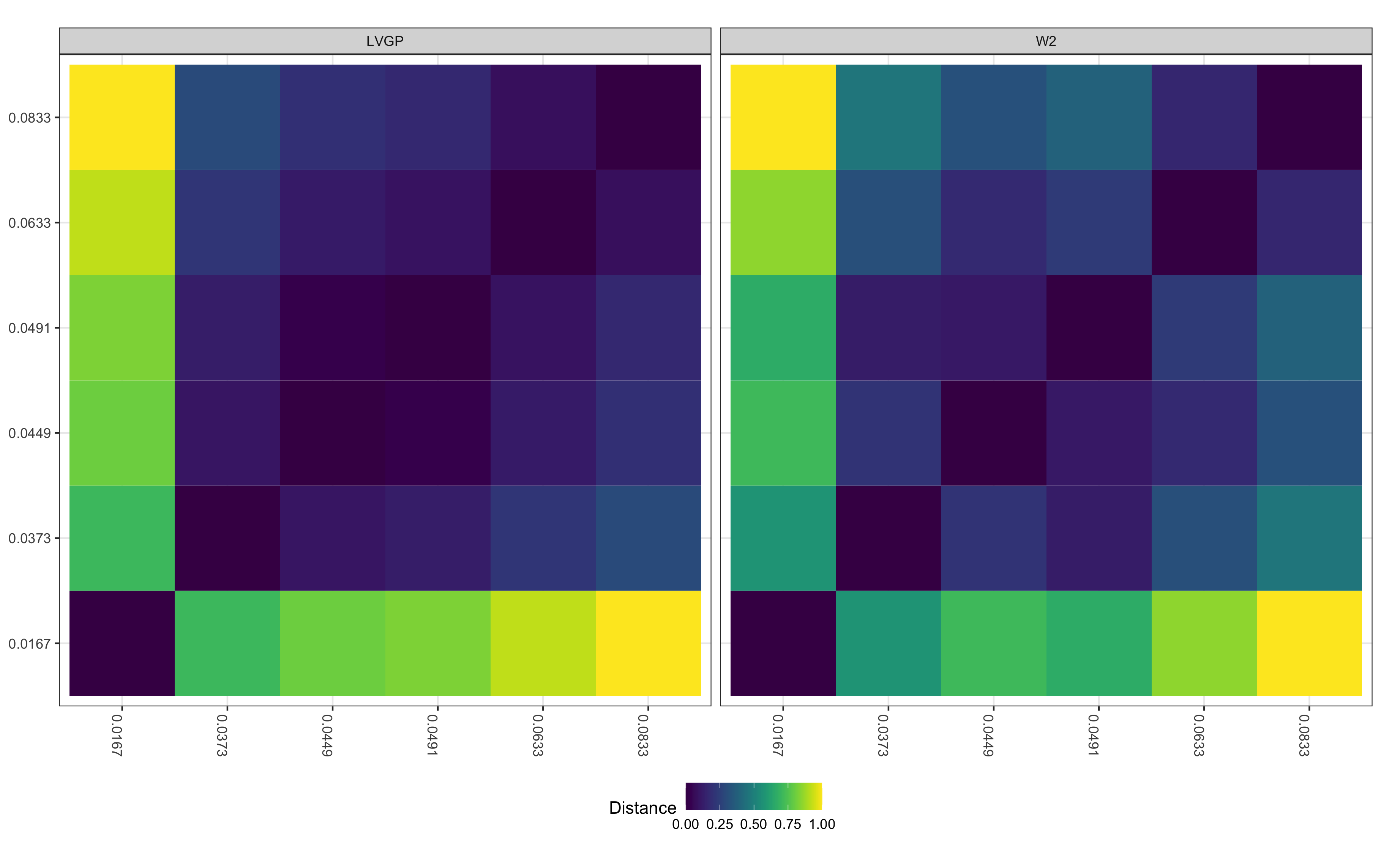} 
\caption{Normalized distance matrix between levels for LVGP and $W_2$ for the beam bending test case.}
\label{fig:beam_distW2}
\end{figure}

The first class of kernels relies on optimal transport distances, such as the Wasserstein distance. These kernels define similarity based on the minimal cost of transporting mass from one distribution to another, incorporating the geometry of the underlying space. We focus here on the Wasserstein distance, which is recalled below.
\begin{definition}[Wasserstein distance]
\label{def:wasserstein}
Let $d \geq 1$ be an integer, $r \geq 1$ be a real number, and $P, Q$ be two probability measures on $\mathbb{R}^d$ having finite moments of order $r$. The $r$-Wasserstein distance is defined as
\begin{equation}
W_r(P, Q) := \bigg( \inf\limits_{\pi \in \Pi(P, Q)} \int\limits_{\mathbb{R}^d\times\mathbb{R}^d} ||\mathbf{x}-\mathbf{y}||^r d\pi(\mathbf{x},\mathbf{y}) \bigg)^{\frac{1}{r}}\,,  
\end{equation}
where $\Pi(P, Q)$ is the set of all probability measures on $\mathbb{R}^d\times\mathbb{R}^d$ whose marginals w.r.t. the first and second variables are respectively $P$ and $Q$ and $||\cdot||$ stands for the Euclidean norm on $\mathbb{R}^d$.
\end{definition}
When dealing with empirical measures, as in our framework, the Wasserstein distance can be computed with linear programming in $\mathcal{O}(n^3 \log(n))$ or accelerated with entropy regularization (e.g. Sinkhorn iterations \citep{sinkhorn}) in $\mathcal{O}(n^2 \log(n))$. A notable exception is when \(d=1\), where the Wassertein distance can be expressed as
\begin{equation*}
\textrm{W}_r(P,Q) = \left( \int_0^1 \vert F_P^{-1}(t) - F_Q^{-1}(t) \vert^r dt \right)^{\frac{1}{r}},
\end{equation*}
with a natural estimator based on the quantile function:
\begin{equation*}
\widehat{\textrm{W}}_r(P,Q) = \left( \frac{1}{Q}\sum_{q=1}^Q  (\hat{F}_P^{-1}(t_q) - \hat{F}_Q^{-1}(t_q))^r \right)^{\frac{1}{r}}
\end{equation*}
where \(t_1,\ldots,t_Q\) is a sequence of equally-spaced points in \([0,1]\). Distance substitution kernels \citep{distance_substitution} then offer a natural way to build kernel functions from the Wasserstein distance. Note that it is not possible to design positive semi-definite kernels by plugging the Wasserstein distance into such kernels when the dimension $d$ of the space $\mathbb{R}^d$ is greater than one \citep{peyre2019}. However, this means that in our setting with only one output, a valid kernel is
\begin{equation*}
k_{W_2}(P,Q) = \exp(- \gamma \, W_2^\beta(P,Q)) 
\end{equation*}
for $P, Q$ two probability measures on $\mathbb{R}$ with finite moments of order $2$, $\beta\in [0,2]$ and $\gamma>0$, see \citet{bachoc2017gaussian}. We will discuss potential extensions for \(d>1\) in Section \ref{sec:extensions}. Returning to the beam bending case, we can compute the $W_2$ distance between empirical measures $\hat{P}^{Y}_{t,u_t}$ for all levels, and use multidimensional scaling to visualize the encodings, see Figure \ref{fig:beam_encodingW2}. Here, we obtain again a latent representation in accordance with LVGP and mean/standard deviation encoding, with a similar distance matrix (Figure \ref{fig:beam_distW2}).

\medskip

Another approach is kernel mean embedding, where a distribution is mapped into a reproducing kernel Hilbert space (RKHS) via its mean element, for some base kernel \(k\). The similarity between two distributions is then measured by the inner product of their embeddings, or alternatively by the squared RKHS distance known as the maximum mean discrepancy (MMD):

\begin{definition}[Maximum Mean Discrepancy \citep{smola2007hilbert}]
Let $X$ and $Y$ be random vectors defined on a topological space $\mathcal{Z}$, with respective Borel probability measures $P$ and $Q$. 
Let $k : \mathcal{Z} \times \mathcal{Z} \rightarrow \mathbb{R}$ be a kernel function and let $\mathcal{H}(k)$ be the associated reproducing kernel Hilbert space. The maximum mean discrepancy between $P$ and $Q$ is defined as
\begin{align*}
    \mathrm{MMD}(P, Q) = \sup_{\|f\|_{\mathcal{H}(k)} \leq 1} |\mathbb{E}_{X \sim P}[f(X)] - \mathbb{E}_{Y\sim Q}[f(Y)]|\,,
\end{align*}
with the following closed-form expression:
\begin{align*}
    \mathrm{MMD}(P, Q)^2 &= \mathbb{E}_{X\sim P,X' \sim P}[k(X,X')] + \mathbb{E}_{Y \sim Q,Y' \sim Q}[k(Y,Y')] \\
    &- 2\mathbb{E}_{X \sim P,Y \sim Q}[k(X,Y)]\,,
\end{align*}
which can be estimated thanks to U- or V-statistics.
\end{definition}

Interestingly, \citet{song2008} showed that
\begin{equation*}
k_{\textrm{MMD}}(P,Q) = \exp(- \gamma \, \textrm{MMD}^2(P,Q)),
\end{equation*}
the distance substitution kernel obtained by plugging the MMD within the Gaussian kernel, is positive semi-definite whatever the dimension of $\mathcal{Z}$.

\begin{remark}
    One may argue that using the MMD kernel involves the choice of a new kernel with additional hyperparameters to tune. Even if rules of thumb have been proposed for the Gaussian kernel \citep{smola2007hilbert}, we choose here the kernel $k(X,X')=\frac{1}{2}\left(\Vert X\Vert + \Vert X'\Vert - \Vert X-X'\Vert\right)$ which does not depend on hyperparameters. Plugging this kernel into the MMD is in fact equivalent to using the energy distance \citep{szekely2013energy}, as was shown by \citet{sejdinovic2013equivalence}.
\end{remark}

\subsection{Auxiliary variables}

Beyond improved expressiveness, distributional encoding also provides a framework for integrating easily valuable information from auxiliary variables. In fact, in engineering applications or material science, qualitative inputs typically represent meaningful categorical differences, such as experimental setups, material types, or population groups, to name a few. This means that for such applications, practitioners usually have access to additional datasets on top of the dataset available for training the GP model. For example, they can consist of:
\begin{itemize}
    \item real experiments at a small scale, which can be cheaper and measure other quantities of interest (this setting is frequent in e.g., material science)
    \item low-fidelity numerical simulations, which only capture global trends as opposed to the training dataset composed of high-fidelity points
    \item other related datasets already publicly available
\end{itemize}
In all cases, as long as a qualitative input from our original problem is also present in such an auxiliary dataset, the observed responses can be collected for each level to build a new distributional encoding. From there, the integration into our framework can be performed in two different ways:
\begin{enumerate}
    \item We can concatenate the auxiliary response with the samples from the training set if the responses correspond to the same quantity of interest
    \item We can create a new virtual input with the auxiliary encoding only 
\end{enumerate}
The latter is particularly suited when several additional sources of information are available, but at the cost of increasing the problem dimension. From a practical viewpoint, this is related to multi-task learning, which we discuss in Section \ref{sec:extensions}. The former preserves the problem dimension and possesses an attractive feature: its ability to handle previously unseen levels of the qualitative input. If auxiliary observations are available for a new level (not present in the training dataset), a distribution encoding can be formed and plugged directly into the kernel, allowing predictions without retraining the GP. This is in contrast to latent variable methods, which require fitting an embedding for each new level.

\begin{remark}
    In the context of low-fidelity numerical simulations, \citet{ginsbourger2013distance} previously proposed to define a kernel between functional inputs via a distance between the outputs obtained from proxy simulations. Our approach to handle auxiliary data is similar in spirit.
\end{remark}

\subsection{Extensions}
\label{sec:extensions}

So far, we have considered distributional encoding in the univariate regression setting only, but it can actually accommodate easily other statistical learning problems such as classification or multi-task learning. We defer to Appendix \ref{sec:gsa} a potential extension for problems with strong feature interactions.

\subsubsection{Classification}

The first extension concerns classification problems, where the response is also a qualitative variable. In this setting, distributional encoding thus boils down to considering the empirical histograms of the output categories for each level of a qualitative input, as illustrated in Table \ref{tab:dencoding_classif}.

\begin{table*}[hbt!]
\centering
        \large
        \begin{minipage}{0.45\linewidth}
            \centering
            
            \begin{tabular}{|c|c|c|c|}
\hline
$\mathbf{X_1}$ & $\mathbf{X_2}$ & $\mathbf{U_1}$ & $\mathbf{Y}$ \\
\hline
0.47 & -1.47 & \textcolor{IndianRed3}{red} &  apple\\
\hline
0.52 & -0.79 & \textcolor{SeaGreen4}{green} &   apple\\
\hline
0.11 & -2.67 & \textcolor{SeaGreen4}{green} &  banana\\
\hline
0.75 & 0.43 & \textcolor{SteelBlue4}{blue} &  orange\\
\hline
0.11 & 1.91 & \textcolor{IndianRed3}{red} &  orange\\
\hline
0.96 & 2.92 & \textcolor{SteelBlue4}{blue} &  banana\\
\hline
0.64 & 0.33 & \textcolor{SteelBlue4}{blue} &  apple \\
\hline
0.01 & 2.14 & \textcolor{IndianRed3}{red} &  banana\\
\hline
0.15 & 1.39 & \textcolor{SeaGreen4}{green} &  orange\\
\hline
0.63 & -1.93 & \textcolor{IndianRed3}{red} &  banana\\
\hline
\end{tabular}
\caption{Original classification dataset.}
\label{tab:dataset_classif}
        \end{minipage}
        \hspace{1cm}
        \begin{minipage}{.45\linewidth}
            \centering
            \begin{tabular}{|c|c|c|c|}
\hline
$\mathbf{X_1}$ & $\mathbf{X_2}$ & $\mathbf{X_3}$ & $\mathbf{Y}$ \\
\hline
0.47 & -1.47 & \includegraphics[width=1cm, height=0.4cm]{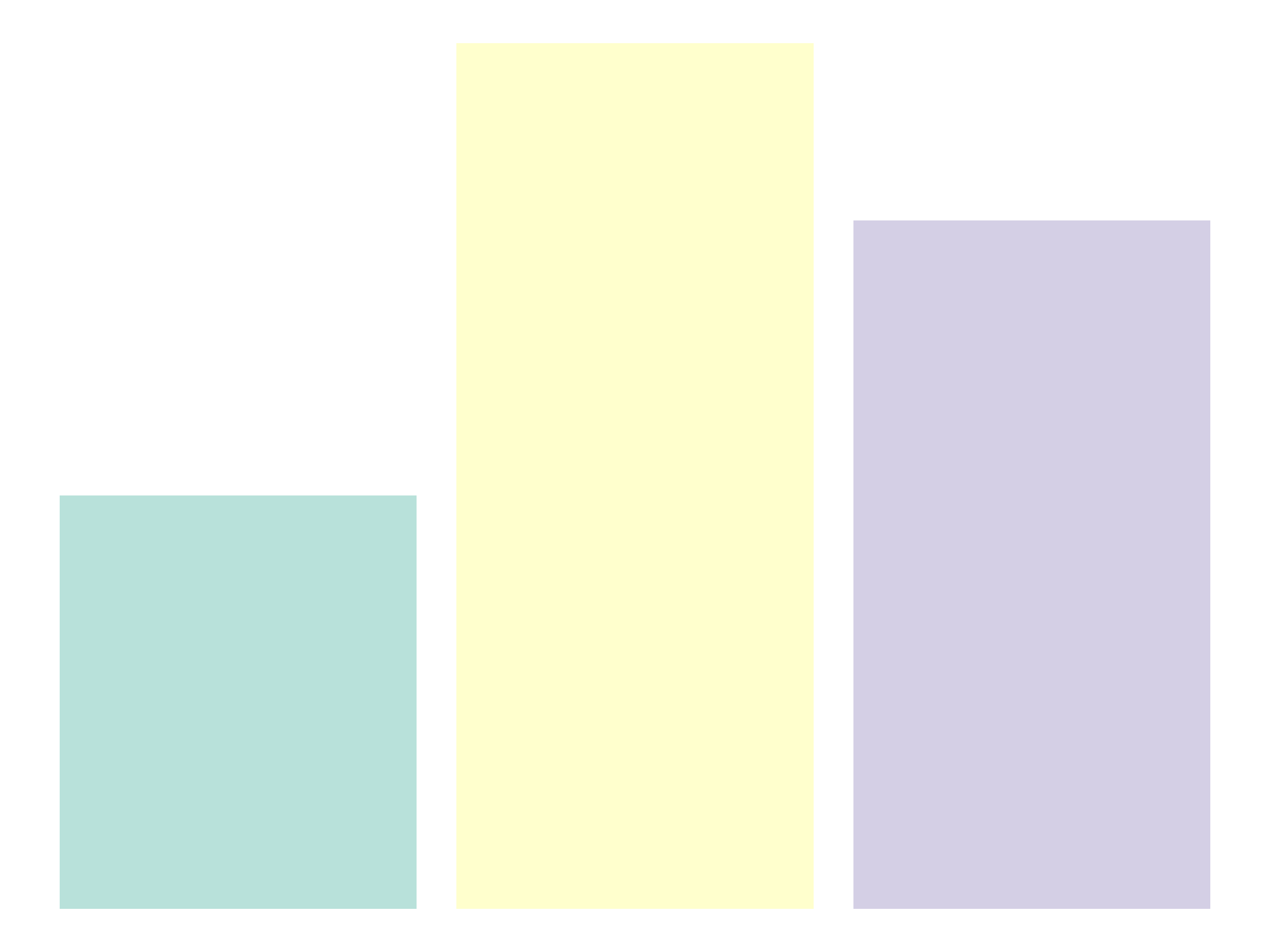} &  apple\\
\hline
0.52 & -0.79 & \includegraphics[width=1cm, height=0.4cm]{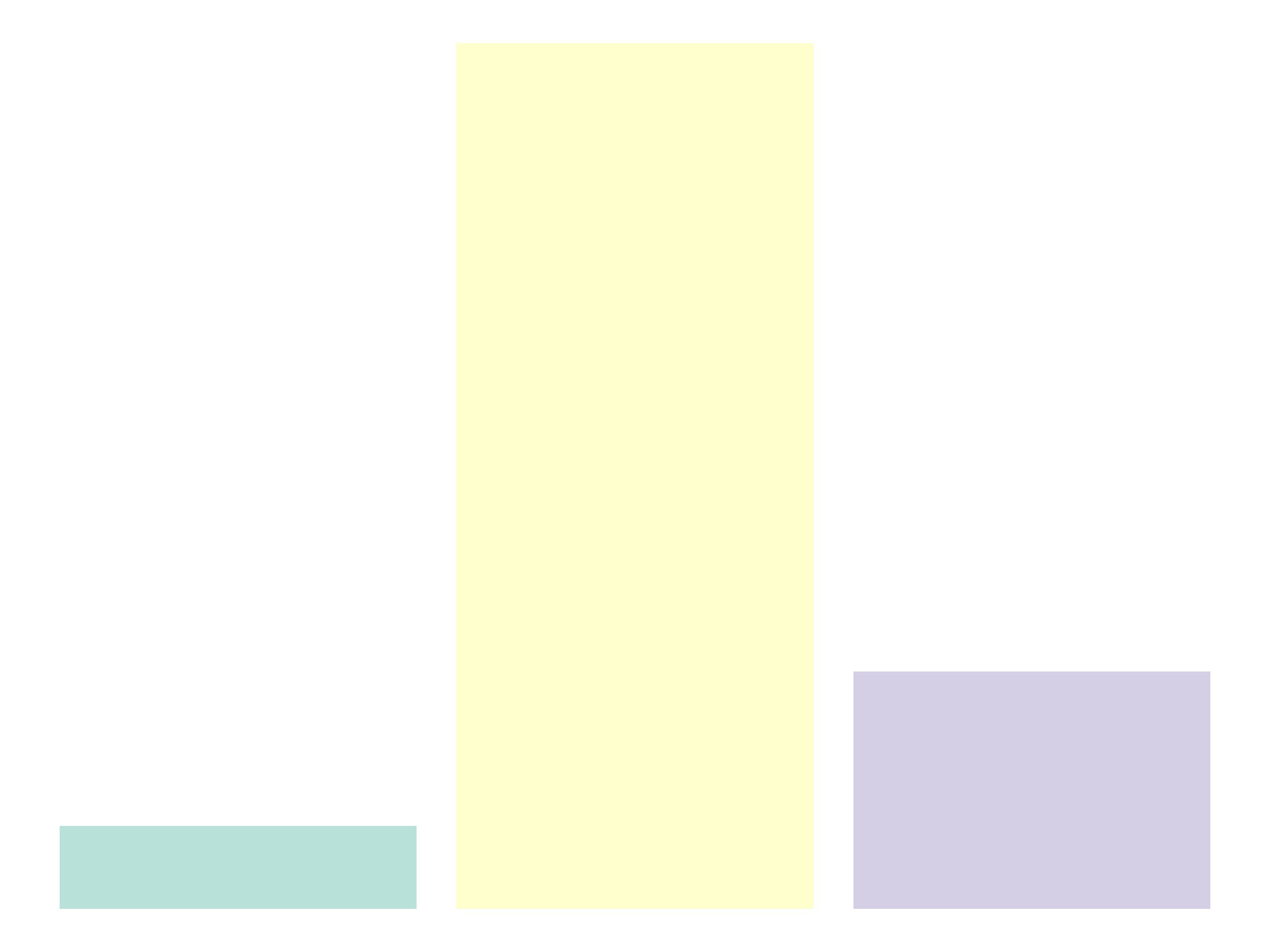} &  apple\\
\hline
0.11 & -2.67 & \includegraphics[width=1cm, height=0.4cm]{d2_hist.png} &  banana\\
\hline
0.75 & 0.43 & \includegraphics[width=1cm, height=0.4cm]{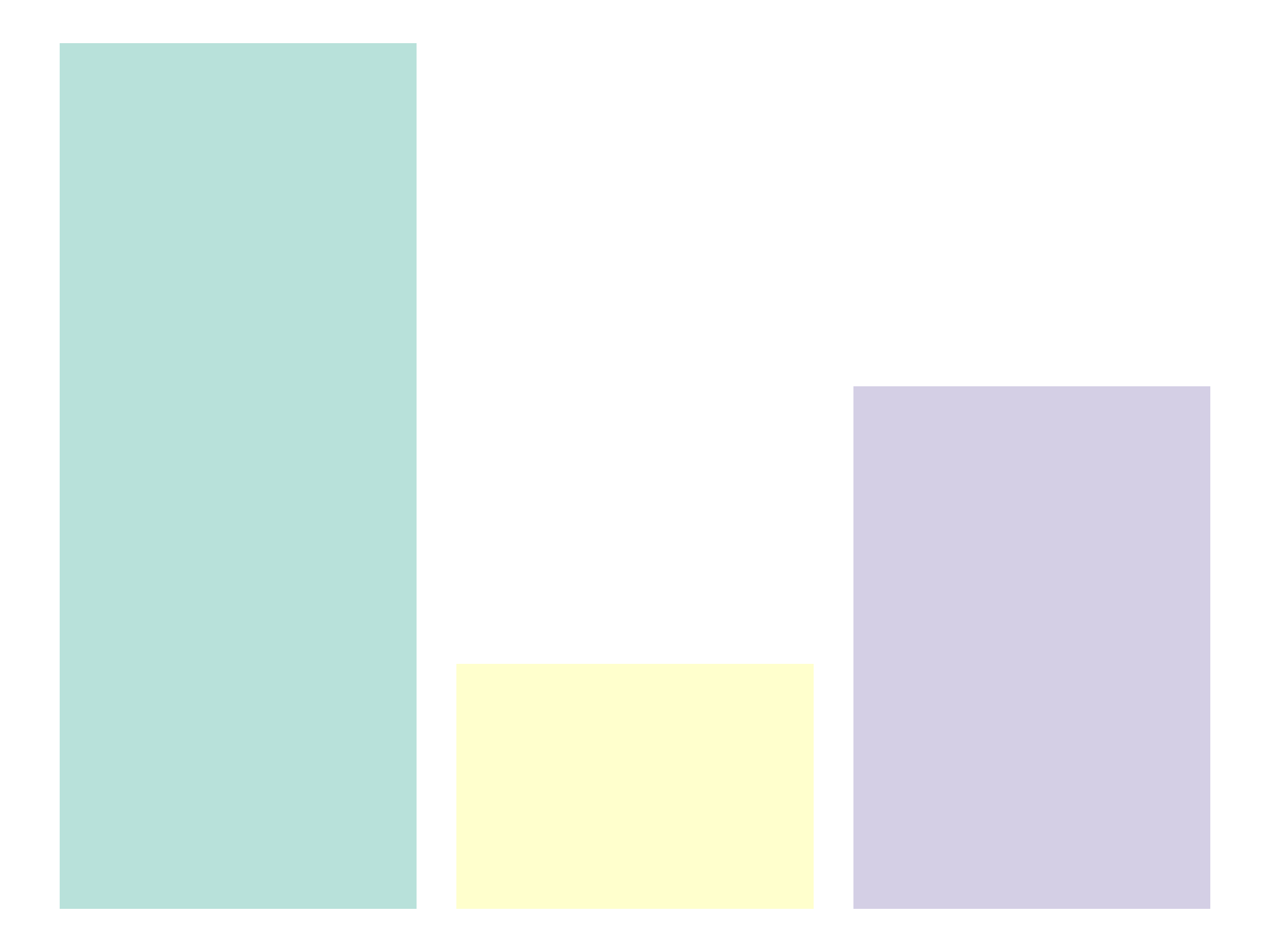} &  orange\\
\hline
0.11 & 1.91 & \includegraphics[width=1cm, height=0.4cm]{d1_hist.png} &   orange\\
\hline
0.96 & 2.92 & \includegraphics[width=1cm, height=0.4cm]{d3_hist.png} &  banana\\
\hline
0.64 & 0.33 & \includegraphics[width=1cm, height=0.4cm]{d3_hist.png} &  apple\\
\hline
0.01 & 2.14 & \includegraphics[width=1cm, height=0.4cm]{d1_hist.png} &  banana\\
\hline
0.15 & 1.39 & \includegraphics[width=1cm, height=0.4cm]{d2_hist.png} &  orange\\
\hline
0.63 & -1.93 & \includegraphics[width=1cm, height=0.4cm]{d1_hist.png} &  banana\\
\hline
\end{tabular}
\caption{Histogram encoding.}
            \label{tab:dencoding_classif}
        \end{minipage} 
    \end{table*}

A key difference is that now the output space does not have a natural metric structure, making $W_2$ or MMD kernels inefficient. Instead, we rely on kernels defined on histograms such as
\begin{equation*}
k_{\textrm{hist}}(H,H') = \exp(-\gamma \, \psi(\mathbf{\theta},\mathbf{\theta}'))
\end{equation*}
where $\mathbf{\theta}=(\theta_m)_{m=1,\ldots,M}$ denotes the frequencies of the $M$ categories from histograms $H$ and $H'$, $\gamma>0$ and $\psi$ is a function to compare frequencies. Standard examples include
\begin{align*}
\psi_{\chi^2}(\mathbf{\theta},\mathbf{\theta}') &= \sum_m \frac{(\theta_m -\theta'_m)^2}{\theta_m +\theta'_m},\\
\psi_{\textrm{TV}}(\mathbf{\theta},\mathbf{\theta}') &= \sum_m \vert \theta_m -\theta'_m\vert,\\
\psi_{H_2}(\mathbf{\theta},\mathbf{\theta}') &= \sum_m (\sqrt{\theta_m} -\sqrt{\theta'_m})^2,
\end{align*}
the $\chi^2$, total-variation and Hellinger distances, see \citet{cuturi2006} for details and illustrations. The rest of the procedure remains unchanged.

\subsubsection{Multi-task learning}

Multi-task learning is a paradigm where multiple related tasks are learned jointly rather than independently. In the context of Gaussian processes, multi-task GPs allow the sharing of statistical strength across tasks by defining a joint prior over all task-specific functions. This is especially advantageous when tasks have limited individual data but exhibit some correlation. A common setup assumes that all tasks share the same input and output spaces, and that the correlation between tasks can be modeled via a task covariance matrix, leading to formulations such as the linear model of coregionalization or intrinsic coregionalization model, see \citet{alvarez2012kernels} for a detailed review.

\medskip

To define distributional encodings in multi-task learning, we now have access to several outputs (one per task). We can then consider either one encoding per task or only one multivariate encoding, as illustrated in Tables \ref{tab:dmoencoding1} and \ref{tab:dmoencoding2}.

\begin{table*}[hbt!]
\centering
        \large
        \begin{minipage}{0.45\linewidth}
            \centering
            \begin{tabular}{|c|c|c|c|c|c|}
\hline
$\mathbf{X_1}$ & $\mathbf{X_2}$ & $\mathbf{X_3}$ & $\mathbf{X_4}$ & $\mathbf{Y_1}$ & $\mathbf{Y_2}$ \\
\hline
0.47 & -1.47 & \includegraphics[width=1cm, height=0.4cm]{d1.png} & \includegraphics[width=1cm, height=0.4cm]{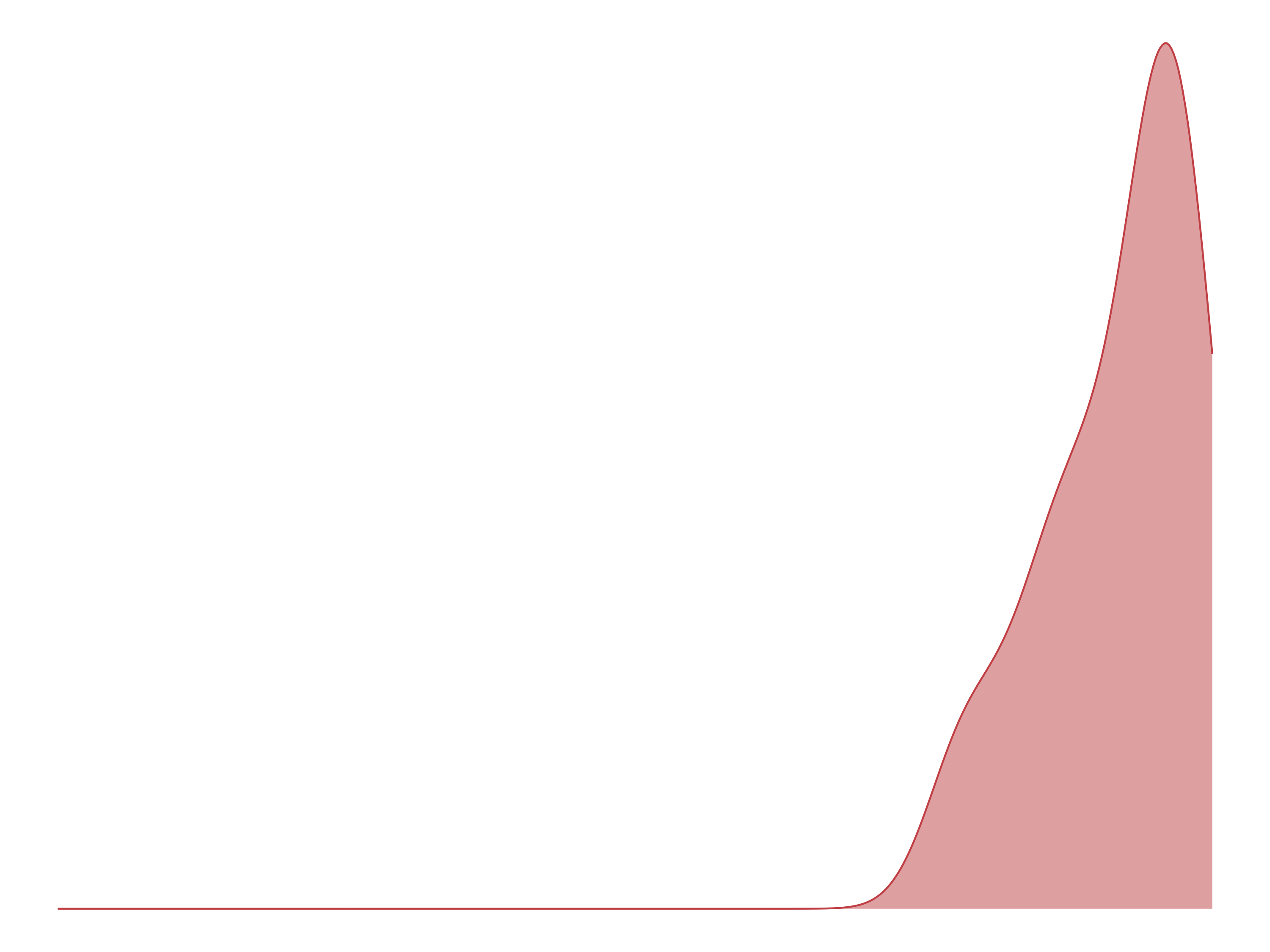} & -1.5 & 5.67\\
\hline
0.52 & -0.79 & \includegraphics[width=1cm, height=0.4cm]{d2.png} & \includegraphics[width=1cm, height=0.4cm]{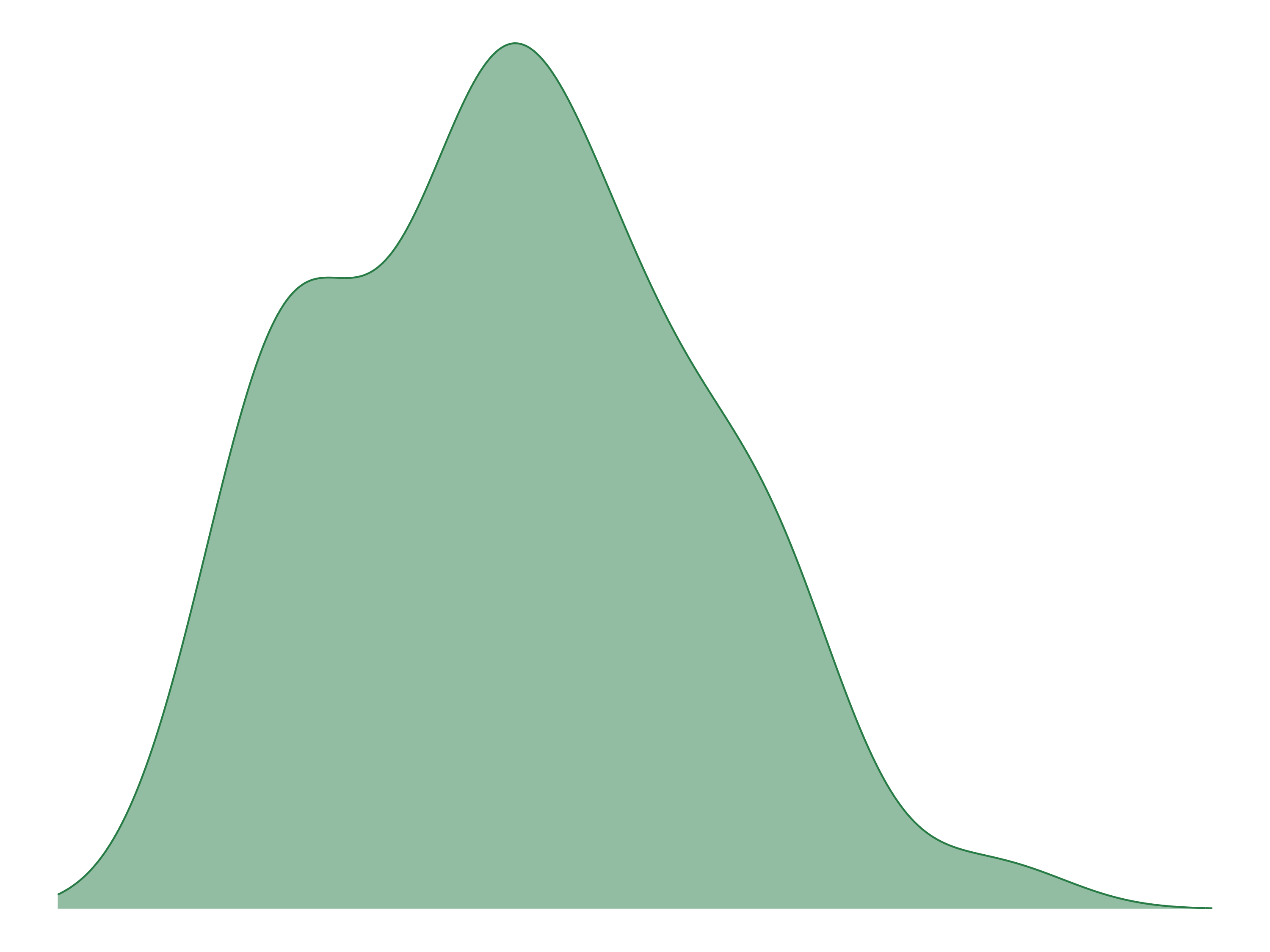} & 0.20 & -0.89\\
\hline
0.11 & -2.67 & \includegraphics[width=1cm, height=0.4cm]{d2.png} & \includegraphics[width=1cm, height=0.4cm]{d2_2.png} & 0.48 & -3.65 \\
\hline
0.75 & 0.43 & \includegraphics[width=1cm, height=0.4cm]{d3.png} & \includegraphics[width=1cm, height=0.4cm]{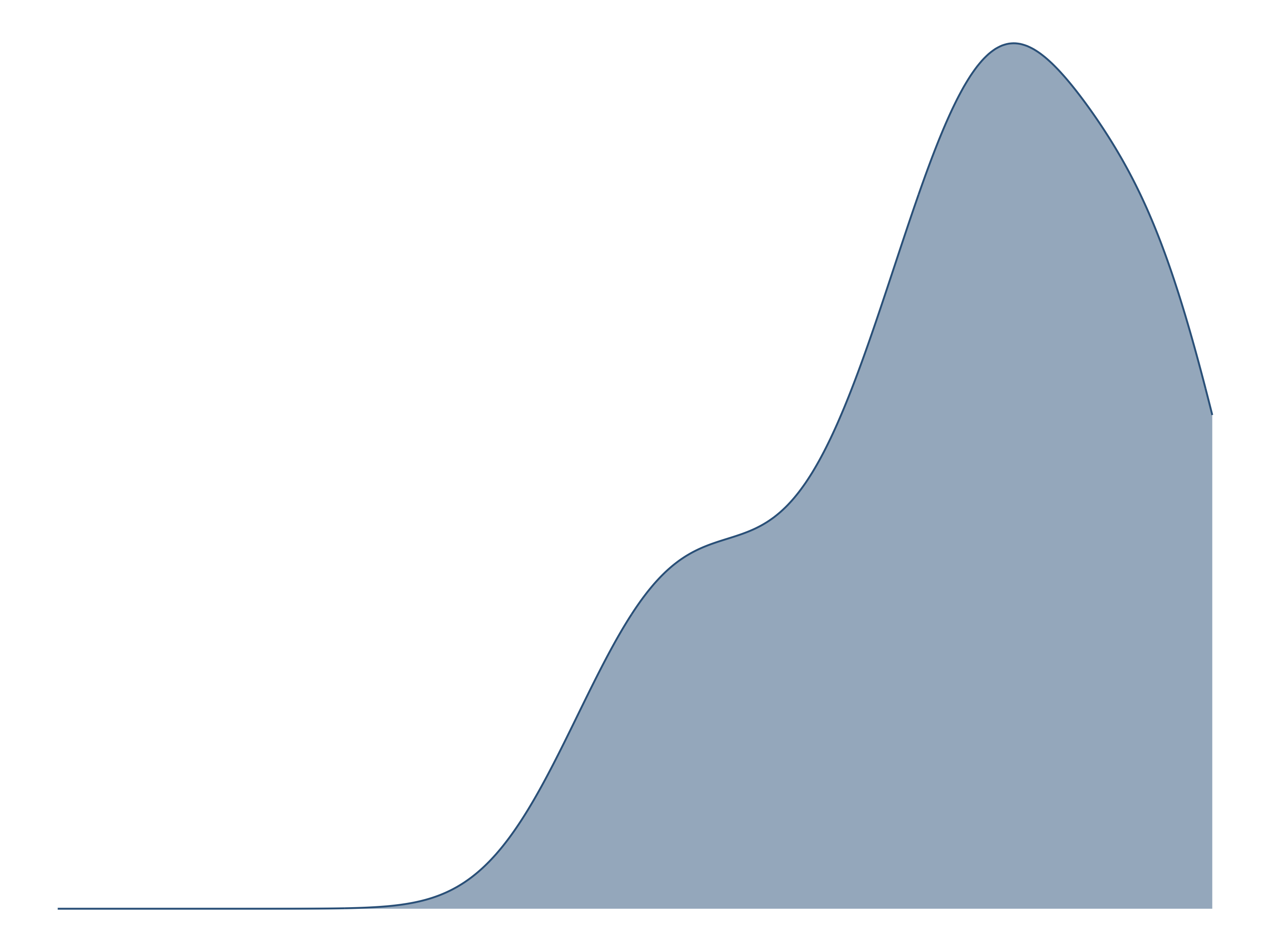} & 1.82 & 7.34\\
\hline
0.11 & 1.91 & \includegraphics[width=1cm, height=0.4cm]{d1.png} & \includegraphics[width=1cm, height=0.4cm]{d1_2.png} & -4.2 & 6.32\\
\hline
0.96 & 2.92 & \includegraphics[width=1cm, height=0.4cm]{d3.png} & \includegraphics[width=1cm, height=0.4cm]{d3_2.png} & 2.34 & 4.28\\
\hline
0.64 & 0.33 & \includegraphics[width=1cm, height=0.4cm]{d3.png} & \includegraphics[width=1cm, height=0.4cm]{d3_2.png} & 4.51 & 10.12\\
\hline
0.01 & 2.14 & \includegraphics[width=1cm, height=0.4cm]{d1.png} & \includegraphics[width=1cm, height=0.4cm]{d1_2.png} & -3.7 & 7.98\\
\hline
0.15 & 1.39 & \includegraphics[width=1cm, height=0.4cm]{d2.png} & \includegraphics[width=1cm, height=0.4cm]{d2_2.png} & 0.86 & 0.73 \\
\hline
0.63 & -1.93 & \includegraphics[width=1cm, height=0.4cm]{d1.png} & \includegraphics[width=1cm, height=0.4cm]{d1_2.png} & -2.9 & 9.21\\
\hline
\end{tabular}
\caption{Multi 1D-Distrib. encoding.}
            \label{tab:dmoencoding1}
        \end{minipage}
        \hfill
        \begin{minipage}{.45\linewidth}
            \centering
            \begin{tabular}{|c|c|c|c|c|}
\hline
$\mathbf{X_1}$ & $\mathbf{X_2}$ & $\mathbf{X_3}$ & $\mathbf{Y_1}$ & $\mathbf{Y_2}$ \\
\hline
0.47 & -1.47 & \includegraphics[width=1cm, height=0.4cm]{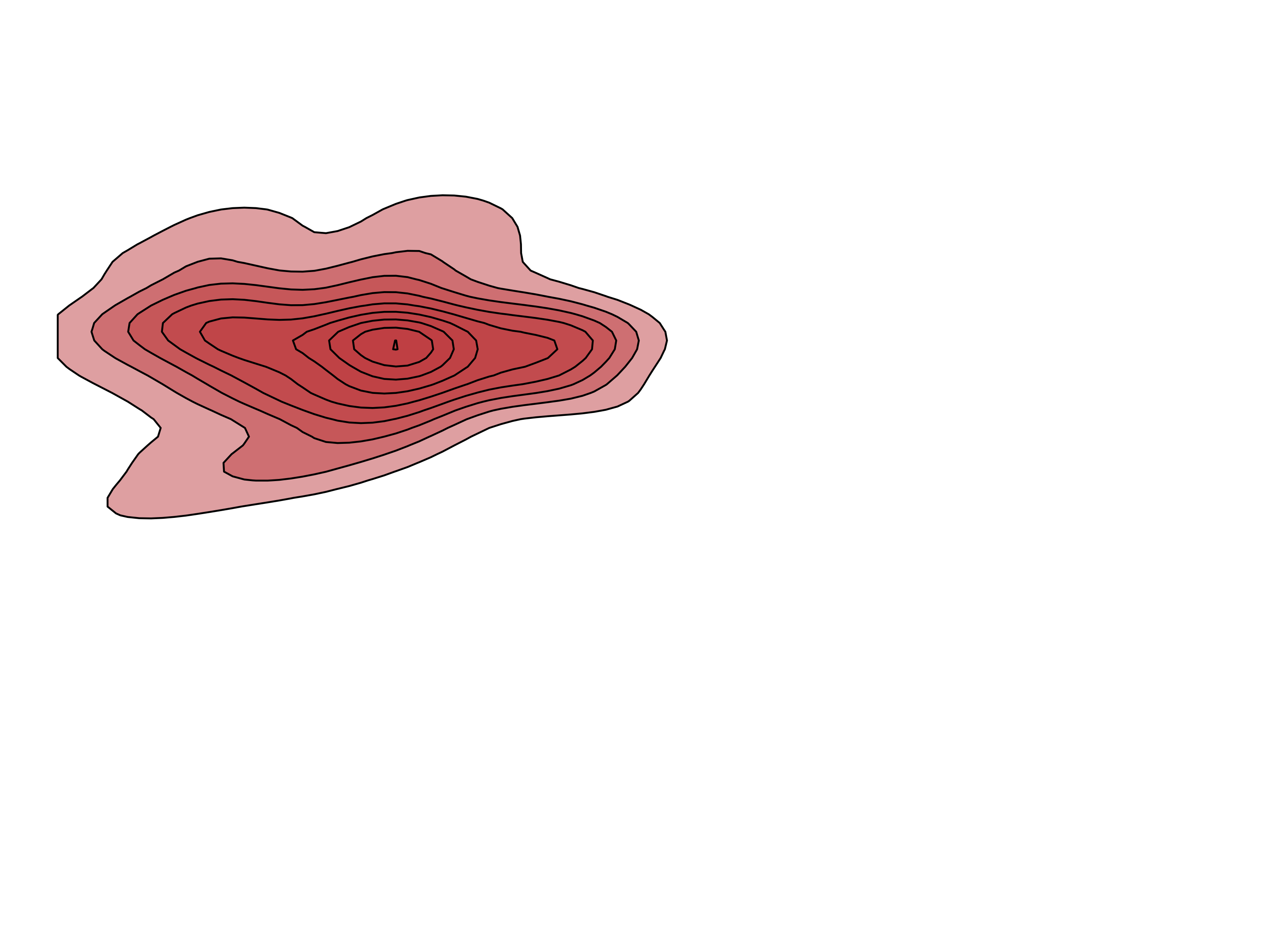} & -1.5 & 5.67\\
\hline
0.52 & -0.79 & \includegraphics[width=1cm, height=0.4cm]{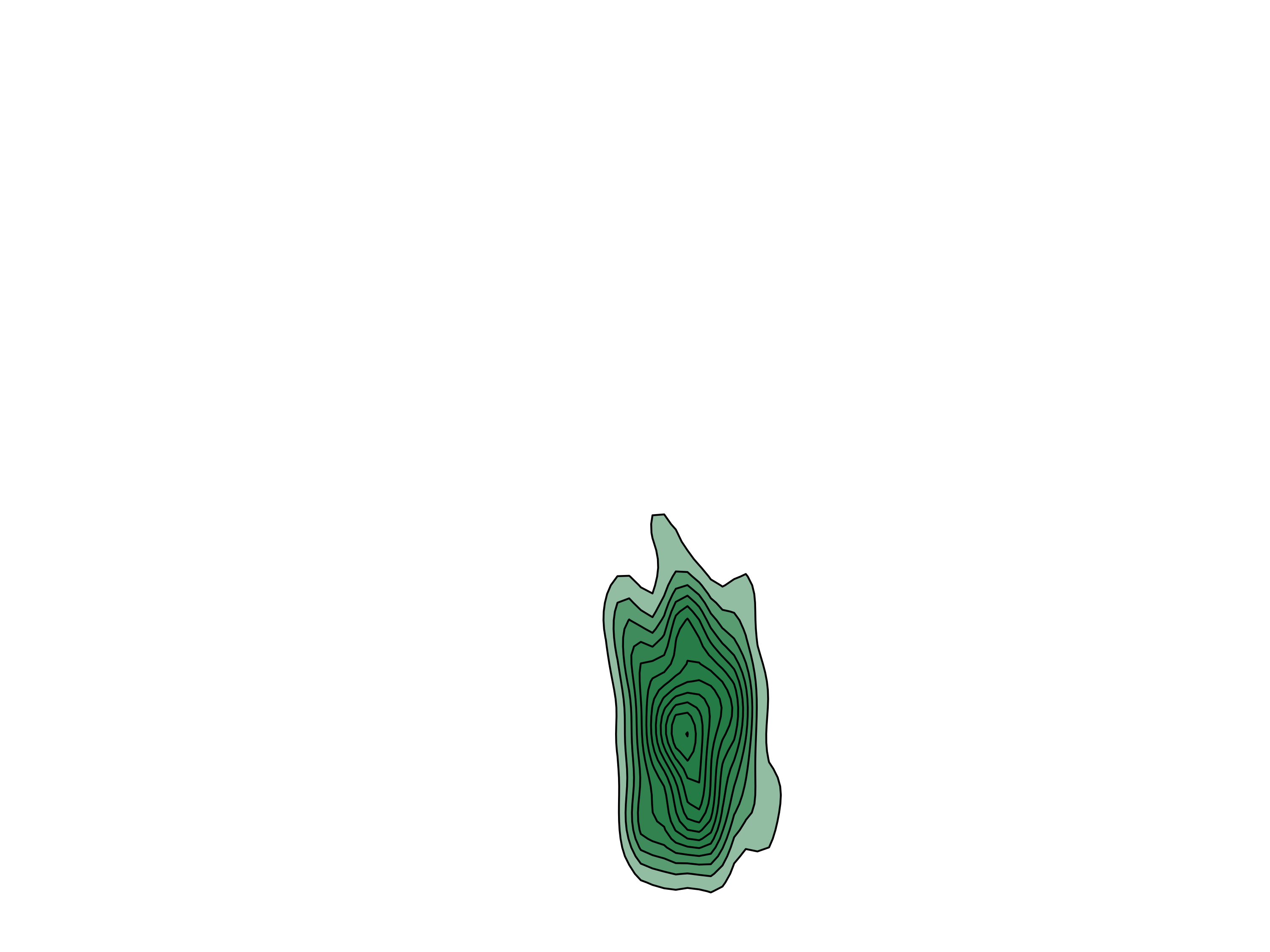} & 0.20 & -0.89\\
\hline
0.11 & -2.67 & \includegraphics[width=1cm, height=0.4cm]{d2_2D.png} & 0.48 & -3.65 \\
\hline
0.75 & 0.43 & \includegraphics[width=1cm, height=0.4cm]{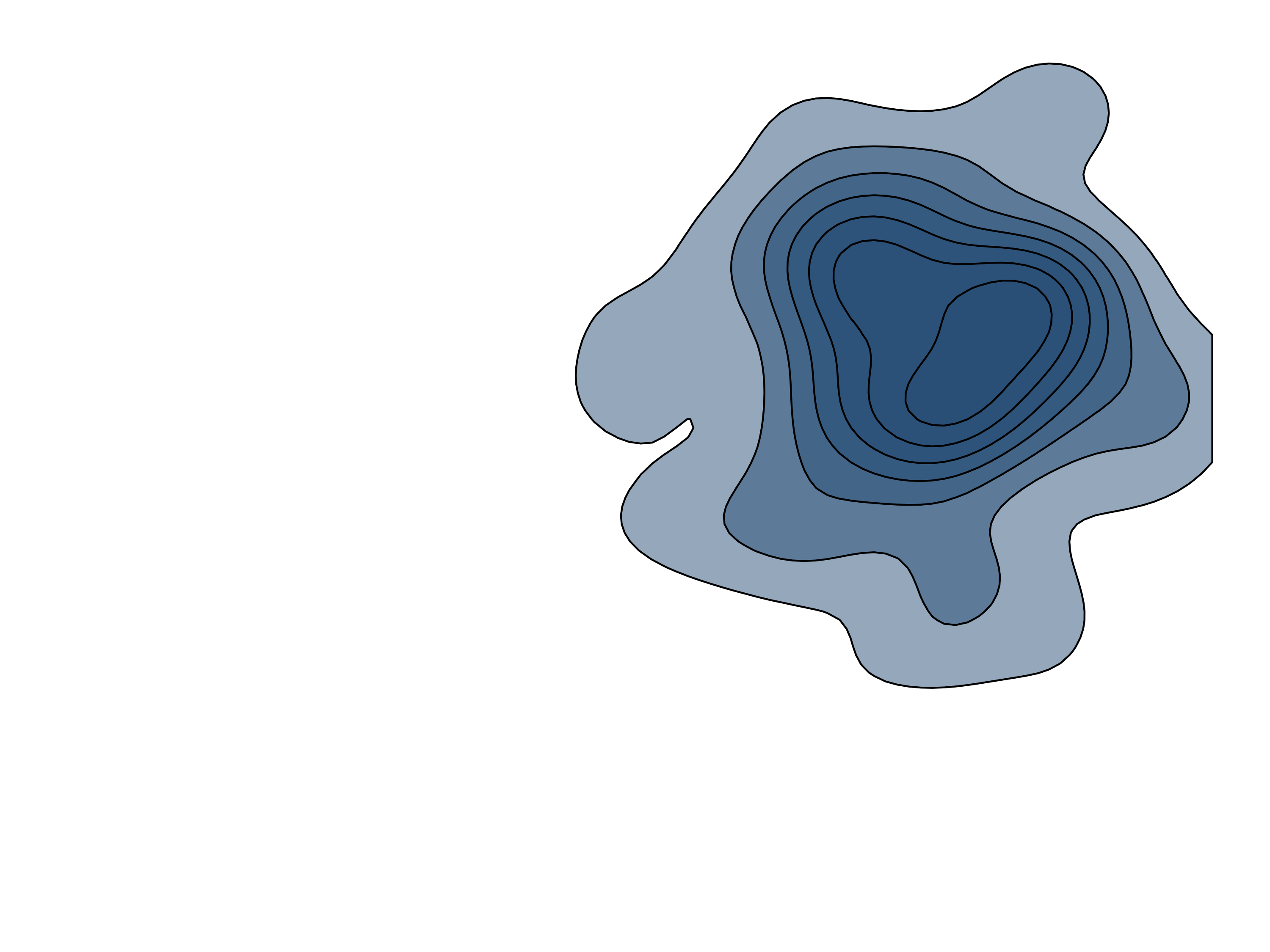} & 1.82 & 7.34\\
\hline
0.11 & 1.91 & \includegraphics[width=1cm, height=0.4cm]{d1_2D.png} & -4.2 & 6.32\\
\hline
0.96 & 2.92 & \includegraphics[width=1cm, height=0.4cm]{d3_2D.png} & 2.34 & 4.28\\
\hline
0.64 & 0.33 & \includegraphics[width=1cm, height=0.4cm]{d3_2D.png} & 4.51 & 10.12\\
\hline
0.01 & 2.14 & \includegraphics[width=1cm, height=0.4cm]{d1_2D.png} & -3.7 & 7.98\\
\hline
0.15 & 1.39 & \includegraphics[width=1cm, height=0.4cm]{d2_2D.png} & 0.86 & 0.73 \\
\hline
0.63 & -1.93 & \includegraphics[width=1cm, height=0.4cm]{d1_2D.png} & -2.9 & 9.21\\
\hline
\end{tabular}
\caption{2D-Distrib. encoding.}
            \label{tab:dmoencoding2}
        \end{minipage} 
    \end{table*}

Considering multiple one-dimensional encodings is a straightforward extension, in the sense that recycling the previously defined kernels on probability distributions is enough. On the contrary, for the multivariate case with $d$ outputs, we define
\begin{eqnarray*}
\hat{P}^{Y_1,\ldots,Y_d}_{t,l}=\frac{1}{N_{tl}} \sum_{i=1}^n \delta_{y_1^{(i)},\ldots,y_d^{(i)}}\mathds{1}_{u_t^{(i)}=l}
\end{eqnarray*}
for $l=1,\ldots,L_j$, the empirical version of the conditional probability distribution 
\begin{eqnarray*}
P(Y_1,\ldots,Y_d \vert u_t=l).
\end{eqnarray*}
Although the MMD kernel is positive semi-definite for any $d$ and can directly handle such multivariate setting, this is not the case for the $W_2$ kernel. To avoid this issue, we rely instead on the sliced Wasserstein distance \citep{sliced_barycenter}, which averages the one-dimensional Wasserstein distances between the distributions projected on the unit sphere. 

\begin{definition}[Sliced Wasserstein distance]
Let $d\geq 1$, $r\geq 1$. The $r$-sliced Wasserstein distance is defined as
\begin{equation} \label{eq:sliced_wasserstein}
SW_r(\mu,\nu) := \bigg(\int\limits_{\mathbb{S}^{d-1}} W_r(\theta^*_\sharp \mu, \theta^*_\sharp \nu)^r d\sigma(\theta) \bigg)^{\frac{1}{r}}\,,
\end{equation}
where $\mathbb{S}^{d-1}$ is the $(d-1)$-dimensional unit sphere, $\sigma$ is the uniform distribution on $\mathbb{S}^{d-1}$,  $\theta^* : \mathbf{x} \mapsto \langle \mathbf{x}, \bs{\theta} \rangle$ the projection function of $\mathbf{x} \in \mathbb{R}^d$ in the direction $\bs{\theta} \in \mathbb{S}^{d-1}$ and $\theta^*_\sharp \mu$ the push-forward measure of $\mu$ by $\theta^*$.
\end{definition}

Interestingly, \citet{meunier2022} showed that the corresponding substitution kernel 
\begin{equation*}
k_{SW_2}(P,Q) = \exp(- \gamma \, SW_2^\beta(P,Q))
\end{equation*}
is positive semi-definite for any \(\beta\in [0,2]\). In addition, estimation of the sliced Wasserstein distance only relies on Monte-Carlo sampling of $R$ directions $\bs{\theta}_1, \ldots, \bs{\theta}_R$ uniformly on $\mathbb{S}^{d-1}$: 

\begin{equation*}
\widehat{SW}_2^2(P,Q) = \frac{1}{R}\sum_{r=1}^R \widehat{W}_2^2(\theta^*_{r,\sharp}P,\theta^*_{r,\sharp}Q),
\end{equation*}
with controlled error bounds \citep{mc_estimate, mc_estimate2}. Improved rate of convergence with control variates was also recently proved \citep{leluc2024sliced}.

\section{Numerical experiments}\label{sec:xp}

In this section, we study distributional encoding on several test
cases representative of engineering applications. In particular, we conduct an extensive comparison with LVGP, the latent variable model proposed by \citet{zhang2020}, on their analytical test cases and a challenging material design problem. We also illustrate two specific extensions of distributional encoding: a multi-output regression problem, and the use of auxiliary data from a low-fidelity simulation model. All the results presented here are reproducible with the R code provided in the supplementary material.

\medskip

For all experiments, LVGP is trained with the R package from the authors \citep{r_lvgp}, while our GP implementation is based on RobustGaSP \citep{rgasp} with a Mat\'{e}rn $5/2$ kernel for all features. We repeat each experiment $50$ times with different random seeds to generate several training and test datasets: for analytical functions, training samples are obtained by a sliced latin hypercube design \citep{ba2015optimal} while test samples come from Monte-Carlo (of size $N_*=3000$), and for real datasets training and test samples are obtained through random splitting. For performance comparison, we use the relative root mean-squared errors on the test set (RRMSE). The computation times of the training phase for all methods are deferred to Appendix \ref{sec:addxp}.

\subsection{Regression problems in engineering}
\label{sec:engxp}

To illustrate the potential of distributional encoding for engineering test cases, we first focus on problems inspired from physics given in \citet{zhang2020}. Among the compared approaches, we do not evaluate covariance parameterization since it was shown to perform worse than LVGP. As for encodings, we investigate mean, mean / standard deviation and distributional encoding with $W_2$ and MMD kernels. The first four problems have the following formulations, with details of their respective inputs available in Table \ref{tab:input_summary}:

\begin{itemize}
    \item Beam bending:
    \[y = \frac{L^3}{3 \times 10^9 h^4 I(t)}.\]
    \item Borehole:
\[y = \frac{2\pi T_u (H_u - H_l)}{
\log \left( \frac{r}{r_w} \right) \left[ 10^{-3} + \frac{2LT_u}{
\log \left( \frac{r}{r_w} \right) r_w^2 K_w
} + \frac{T_u}{T_l} \right]}.\]
    \item OTL circuit:
\[y = \frac{B (V_{b1} + 0.74)(R_{c2} + 9)}{B(R_{c2} + 9) + R_f}
+ \frac{11.35 R_f}{B(R_{c2} + 9) + R_f}
+ \frac{0.74 B R_f}{R_{c1}}\]
where
\[V_{b1} = \frac{12 R_{b2}}{R_{b1} + R_{b2}}.\]
    \item Piston:
\[y = 2\pi \sqrt{\frac{M}{
k + \frac{S^2 P_0 V_0}{V^2} \frac{T_a}{T_0}
}}\]
where
\begin{eqnarray*}
    V &= \frac{S}{2k} \left( A + \sqrt{A^2 + \frac{4k P_0 V_0 T_0}{T_a}} \right),\\
    A &= \frac{P_0 R + 19.62 R - k V_0}{S}.
\end{eqnarray*}
\end{itemize}

\begin{table}[hbt!]
\centering\small
\begin{tabular}{@{}p{2.6cm}p{6.1cm}p{5.6cm}p{1cm}@{}}
\toprule
\textbf{Test case} & \textbf{Continuous inputs} & \textbf{Qualitative inputs \& levels} & $n$\\ \midrule
Beam bending & $L\in[10,20],\;h\in[1,2]$  & $I$\,: \{0.0491, 0.0833, 0.0449, & 90 \\
&   & 0.0633, 0.0373, 0.0167\} & \\[8pt]
Borehole & $r\in[100,50000],\;H_u \in[990, 1110]$  & $r_w$: \{0.05, 0.10, 0.15\} & 180\\
 & $T_{u}\in[63.07,115.6],\;T_{l}\in[63.1, 116]$  & $H_l$: \{700, 740, 780, 820\} & \\
 & $L \in[1120,1680],\;K_{w}\in[9855, 12045]$  &  & \\[4pt]
OTL circuit & $R_{b1}\in[50,150],\;R_{b2}\in[25,70]$  & $R_f$: \{0.5, 1.2, 2.1, 2.9\} & 120\\
 & $R_{c1}\in[1.2,2.5],\;R_{c2}\in[0.25,1.20]$  & $B$: \{50, 100, 150, 200, 250, 300\}  & \\[4pt]
Piston & $R\in[30,60],\;S\in[0.005,0.020]$  & $P_0$: \{9000, 10000, 11000\} & 225\\
 & $V_0\in[0.002,0.010]$ & $k$: \{1000, 2000, 3000, 4000, 5000\} & \\
 & $T_a\in[290,296],\;T_0\in[340,360]$  & & \\
\bottomrule
\end{tabular}
\caption{Input specification for the four engineering test cases.\label{tab:input_summary}}
\end{table}

\begin{figure}[hbt!]

\begin{subfigure}{.475\linewidth}
  \includegraphics[width=\linewidth]{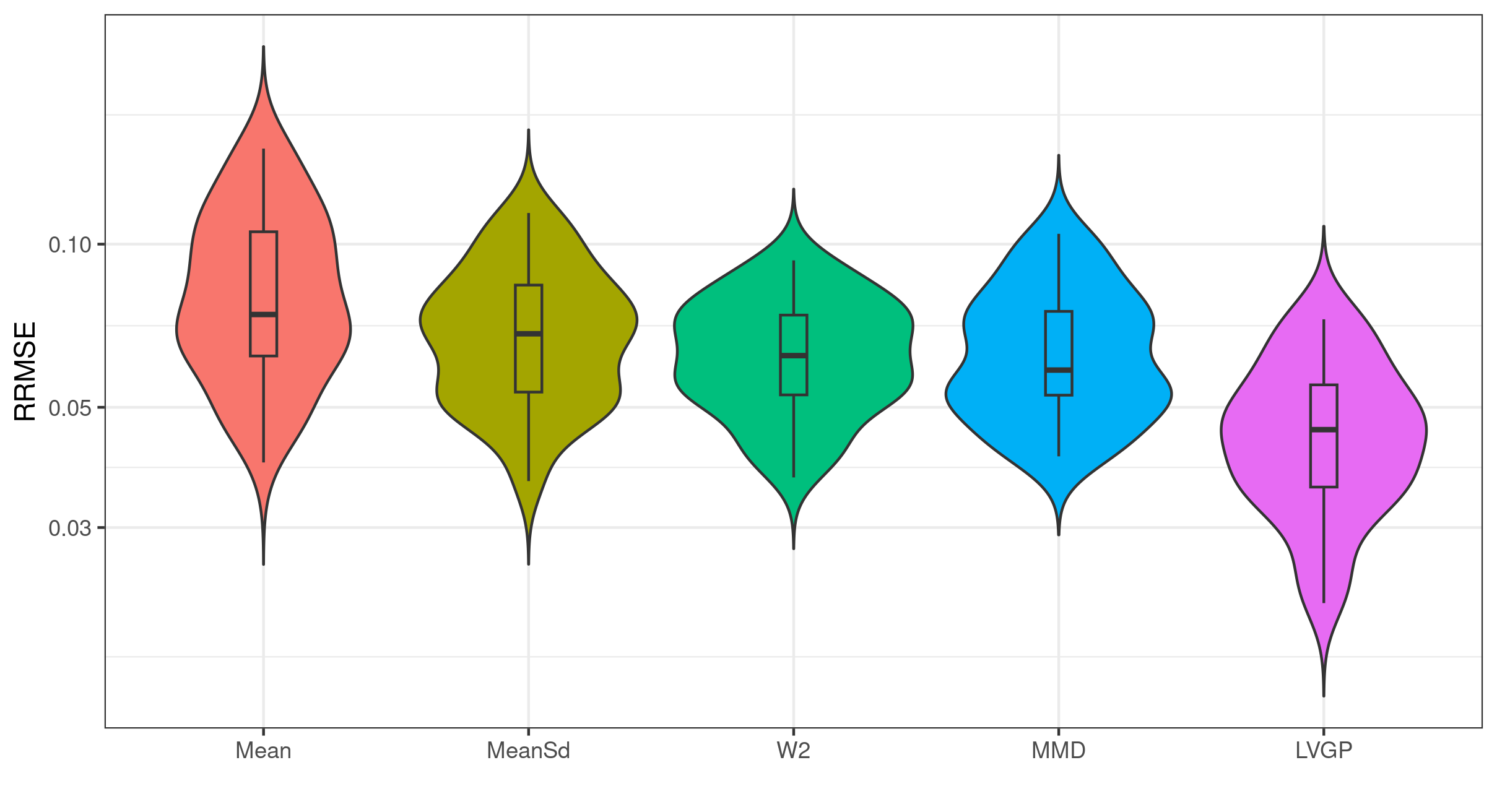}
  \caption{Beam bending.}
  \label{fig:bench_beam}
\end{subfigure}\hfill 
~ 
\begin{subfigure}{.475\linewidth}
  \includegraphics[width=\linewidth]{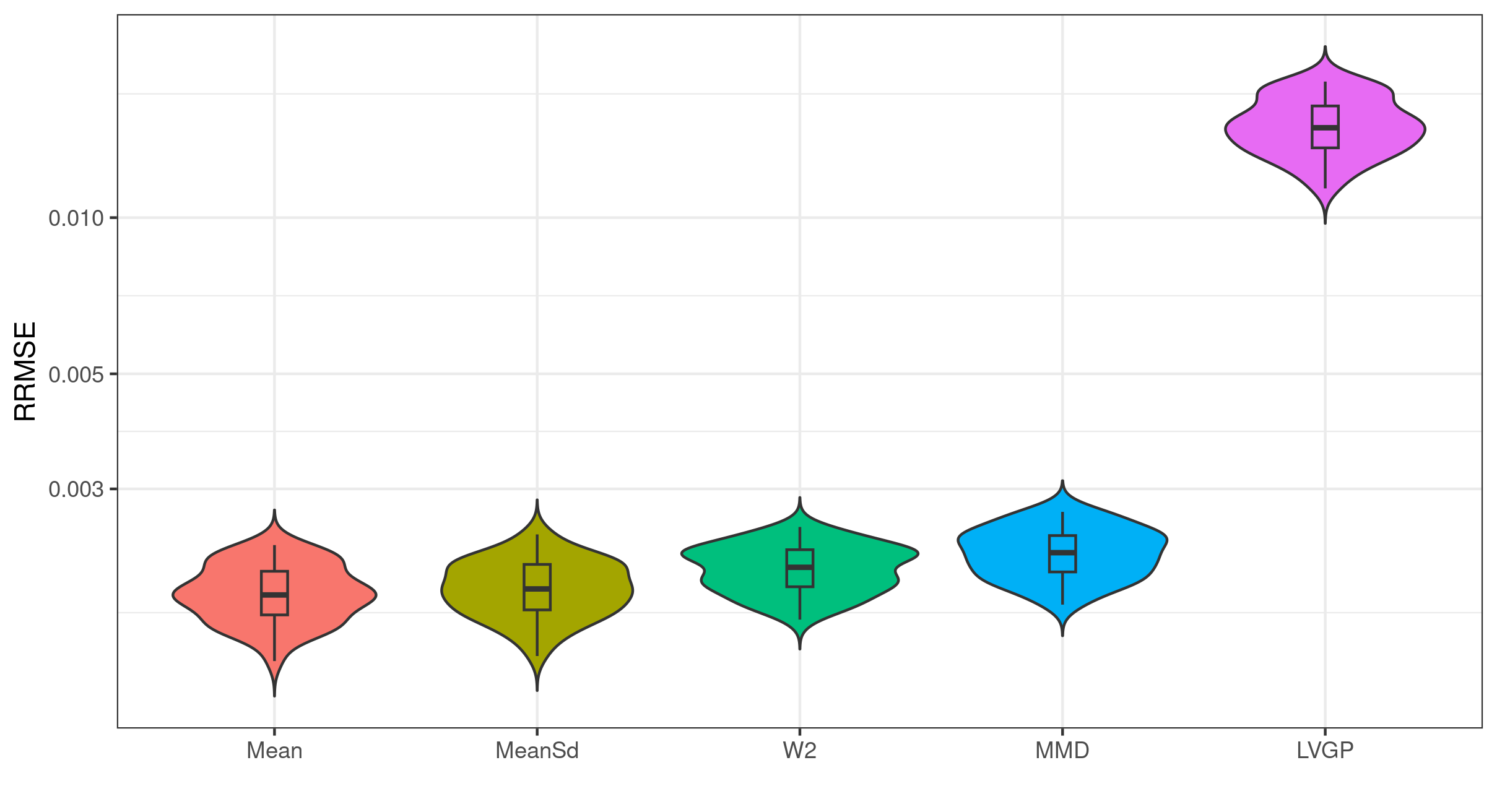}
  \caption{Borehole.}
  \label{fig:bench_borehole}
\end{subfigure}

\medskip 
\begin{subfigure}{.475\linewidth}
  \includegraphics[width=\linewidth]{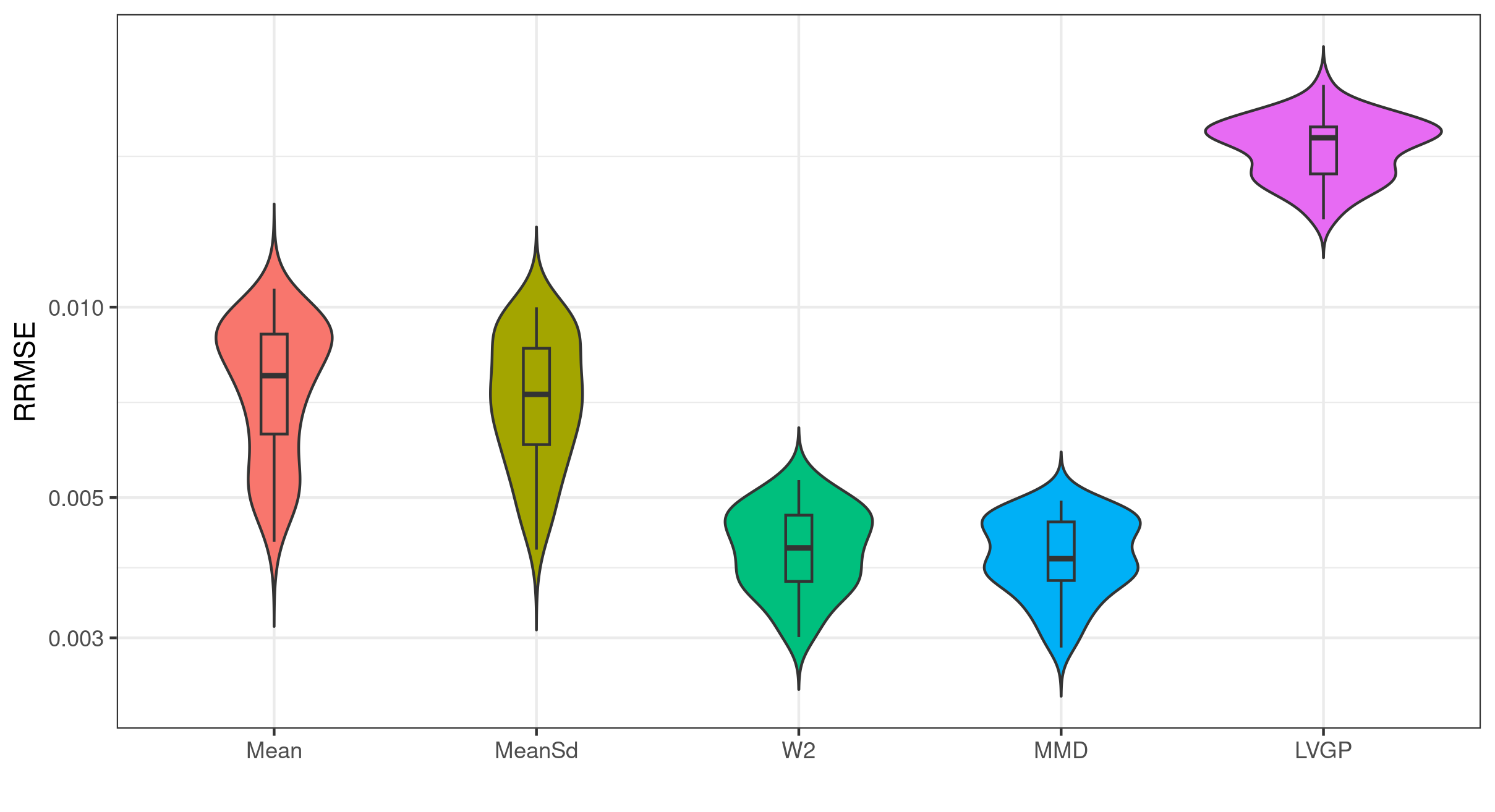}
  \caption{OTL.}
  \label{fig:bench_otl}
\end{subfigure}\hfill 
\begin{subfigure}{.475\linewidth}
  \includegraphics[width=\linewidth]{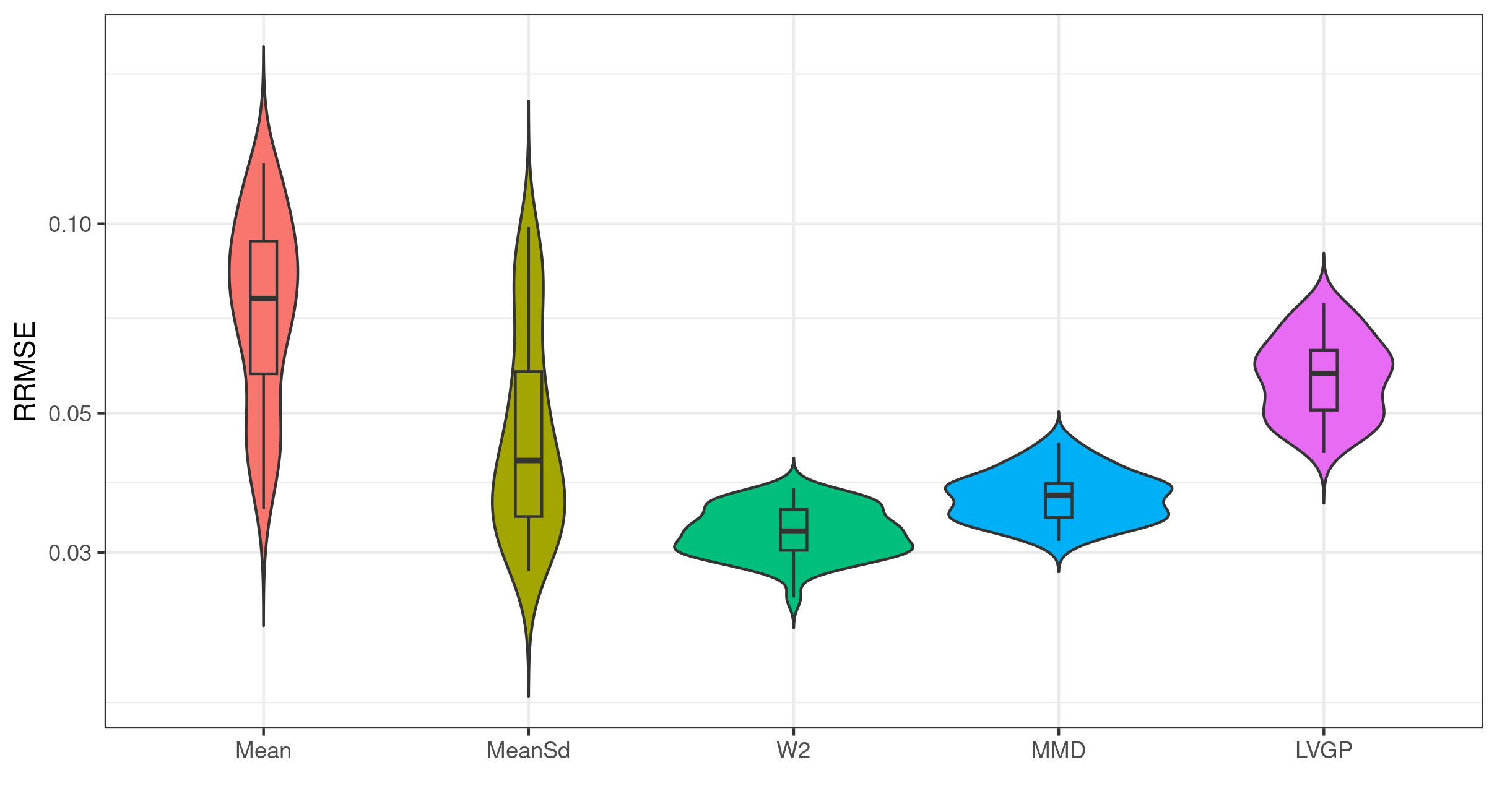}
  \caption{Piston.}
  \label{fig:bench_piston}
\end{subfigure}

\caption{RRMSE on the four engineering test cases with 50 replications.}
\label{fig:benchmark}
\end{figure}

Results for these four problems are displayed in Figure \ref{fig:benchmark}. Except for Beam bending where LVGP has better predictive performance (Figure \ref{fig:bench_beam}), weakly supervised encodings usually have lower RRMSE. Naive mean and mean / standard deviation encodings even yield the best performance for Borehole (Figure \ref{fig:bench_borehole}), while $W_2$ and MMD kernels outperform all methods on OTL and Piston (Figures \ref{fig:bench_otl} and \ref{fig:bench_piston}). Note that aside from Borehole, our LVGP results agree with \citet{zhang2020}.

\begin{figure}[hbt!]
\includegraphics[width=0.95\linewidth]{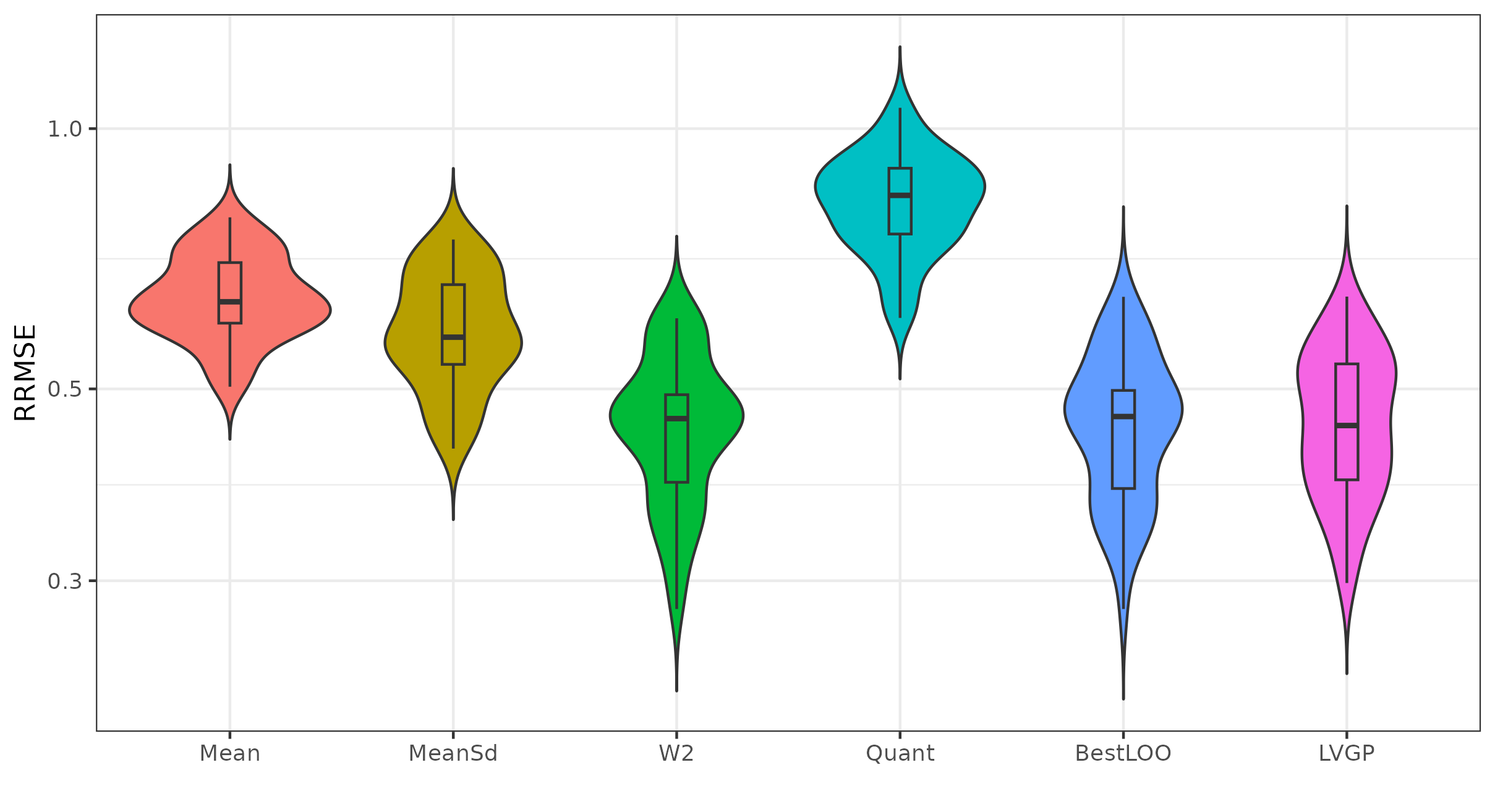}
\caption{RRMSE on the material design test case with 50 replications.}
\label{fig:material}
\end{figure}

A real application from material design \citep{balachandran2016adaptive} was also investigated by \citet{zhang2020}. It consists in predicting the simulated shear modulus of a compound material from its characteristics: choice of M atom among \{Sc, Ti, V, Cr, Zr, Nb, Mo, Hf, Ta, W\}, A atom among \{C, N\}, X atom among \{Al, Si, P, S, Ga, Ge, As, Cd, In, Sn, Tl, Pb\}, three orbital radii for M and two for A and X. There are thus 7 quantitative inputs and 3 qualitative features with 10, 2, and 12 levels, respectively. What makes this problem challenging is that the entire dataset only contains $223$ samples, $200$ of them being used for training and the remaining $23$ for testing, while the number of levels is quite high. Results are reported in Figure \ref{fig:material}, where this time we also evaluate GP with only quantitative inputs (Quant) as a baseline. We observe that mean and mean / standard deviation have interesting predictive performance, between low-performance ones (quantitative) and high-performance ones (LVGP and distributional encoding). Interestingly, distributional encoding is on par with LVGP, but with a much smaller computational cost (see Appendix \ref{sec:addxp} for an illustration). We also leverage fast leave-one-out formulas for GP \citep{dubrule1983cross,ginsbourger2025fast} to rapidly explore all possible combinations of encodings and select the best one for each repetition ("BestLOO" in Figure \ref{fig:material}), see also Appendix \ref{sec:addxp} for a discussion.

\subsection{Multi-task learning problem}

We now consider a simple instance of a multi-task learning problem, multi-output regression, where for each feature sample we observe all outputs. To this end, we use a variant of the Borehole test case proposed in the context of multi-fidelity regression by \citet{xiong2013sequential}:
\[y = \frac{10 T_u (H_u - H_l)}{
\log \left( \frac{r}{r_w} \right) \left[ 1.5\times 10^{-3} + \frac{2LT_u}{
\log \left( \frac{r}{r_w} \right) r_w^2 K_w
} + \frac{T_u}{T_l} \right]}.
\]
Since we expect to benefit more from multivariate encoding in a small data regime, we generate a dataset of size $n=24$ samples only, for which we compute two outputs corresponding to the initial Borehole function and the low-fidelity one above. 

\bigskip

\begin{minipage}{\textwidth}
  \begin{minipage}[hbt!]{0.49\textwidth}
    \centering
    \includegraphics[width=\textwidth]{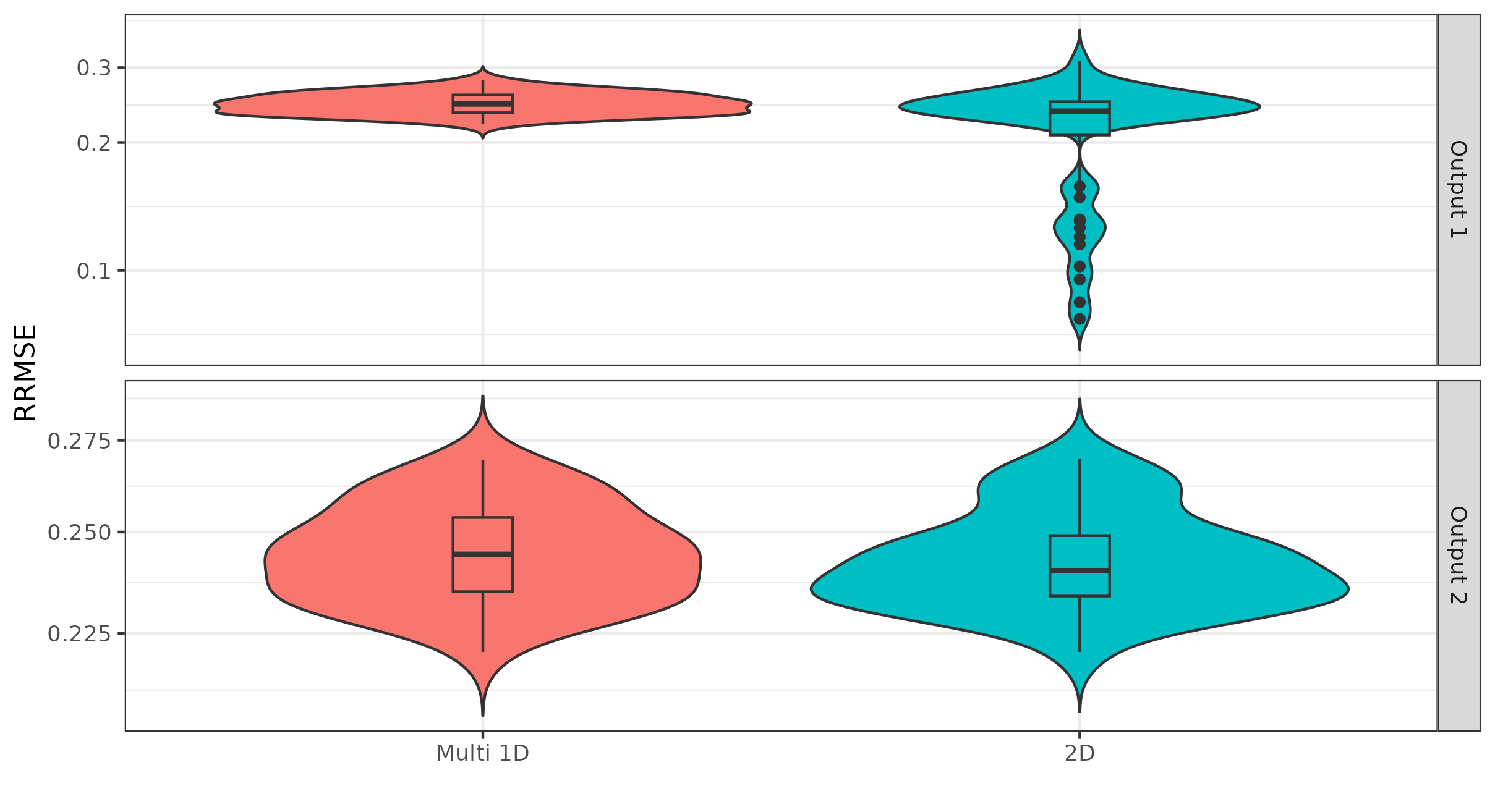}
    \captionof{figure}{RRMSE on the multi-output Borehole test case with 50 replications.}
    \label{fig:borehole_multi}
  \end{minipage}
  \hfill
  \begin{minipage}[hbt!]{0.49\textwidth}
    \centering
    \begin{tabular}{l|c|c}\hline
      & Multi 1D & 2D \\ \hline
      Output 1 & $0.215$ ($0.06$) & $0.21$ ($0.06$) \\ \hline
      Output 2 & $0.221$ ($0.06$) & $0.215$ ($0.06$) \\ \hline
      \end{tabular}
      \captionof{table}{Mean and standard deviation of RRMSE on the multi-output Borehole test case with 50 replications.}
      \label{tab:borehole_multi}
    \end{minipage}
  \end{minipage}

\bigskip

In Figure \ref{fig:borehole_multi}, we compare distributional encoding with the MMD kernel when the outputs are trained independently versus when the joint distribution of the outputs is used for the encoding. The latter exhibits smaller RRMSE consistently, which illustrates the benefits of distribution encoding for multi-task problems, see Table \ref{tab:borehole_multi}.

\subsection{Auxiliary data}

Our last example demonstrates how auxiliary data from a related dataset can help improve GP predictions when used in conjunction with distributional encoding. We place ourselves in a typical computer experiment scenario, where we have a small dataset of high-fidelity samples (Borehole test case with $n=60$) but also a larger dataset of low-fidelity samples of size $180$ according to the model introduced in the previous section.

\medskip

\begin{figure}[hbt!]
\includegraphics[width=0.95\linewidth]{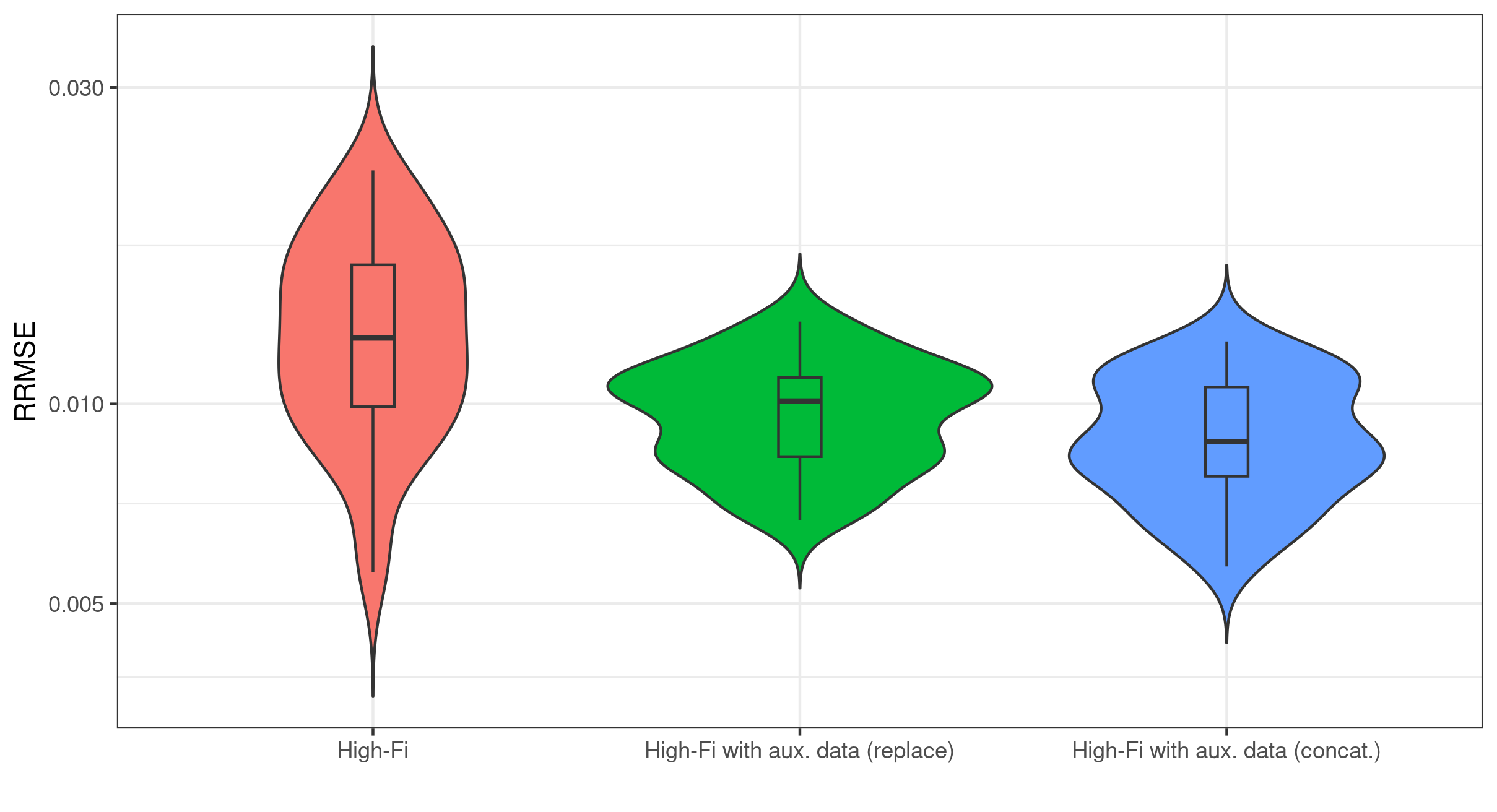}
\caption{RRMSE on the multi-fidelity Borehole test case with 50 replications.}
\label{fig:borehole_multifi}
\end{figure}

We compare in Figure \ref{fig:borehole_multifi} the RRMSE of a model trained with the high-fidelity samples only (distributional encoding and GP training) with a model which only uses low-fidelity samples for encoding -- "replace"-- and one which concatenates high- and low-fidelity samples for encoding --"concat"-- (GP training is still performed on high-fidelity samples only). This example clearly shows that accounting for auxiliary data, which are often available in engineering applications, is quite advantageous.

\section{Conclusion}

In this paper, we introduced the concept of distributional encoding for handling qualitative inputs inside Gaussian process regression. This extension of traditional target encoding is based on recent advances on positive semi-definite kernels for probability distributions. The GP inference, prediction, and hyperparameter optimization procedures remain unchanged, which allows these kernels to be seamlessly integrated into existing GP toolkits. In addition, our approach generalizes easily to classification problems and multi-task learning, and also benefits from auxiliary data when available. 

\medskip

However, there is still room for improvement in the theoretical understanding of theses kernels, including their universality, convergence under empirical approximation, and generalization to sparse or high-dimensional auxiliary data. In addition, hybrid models that combine distributional and latent embeddings could offer the best of both worlds: data-driven flexibility and compact representation. As a simple illustration, latent variable optimization may be initialized with distributional encodings after multi-dimensional scaling.

\bibliography{biblio}
\bibliographystyle{agsm}

\newpage

 \appendix

\section{Links with sensitivity analysis, interactions}
\label{sec:gsa}

\begin{figure}[hbt!]
\centering           
\includegraphics[width=0.8\textwidth]{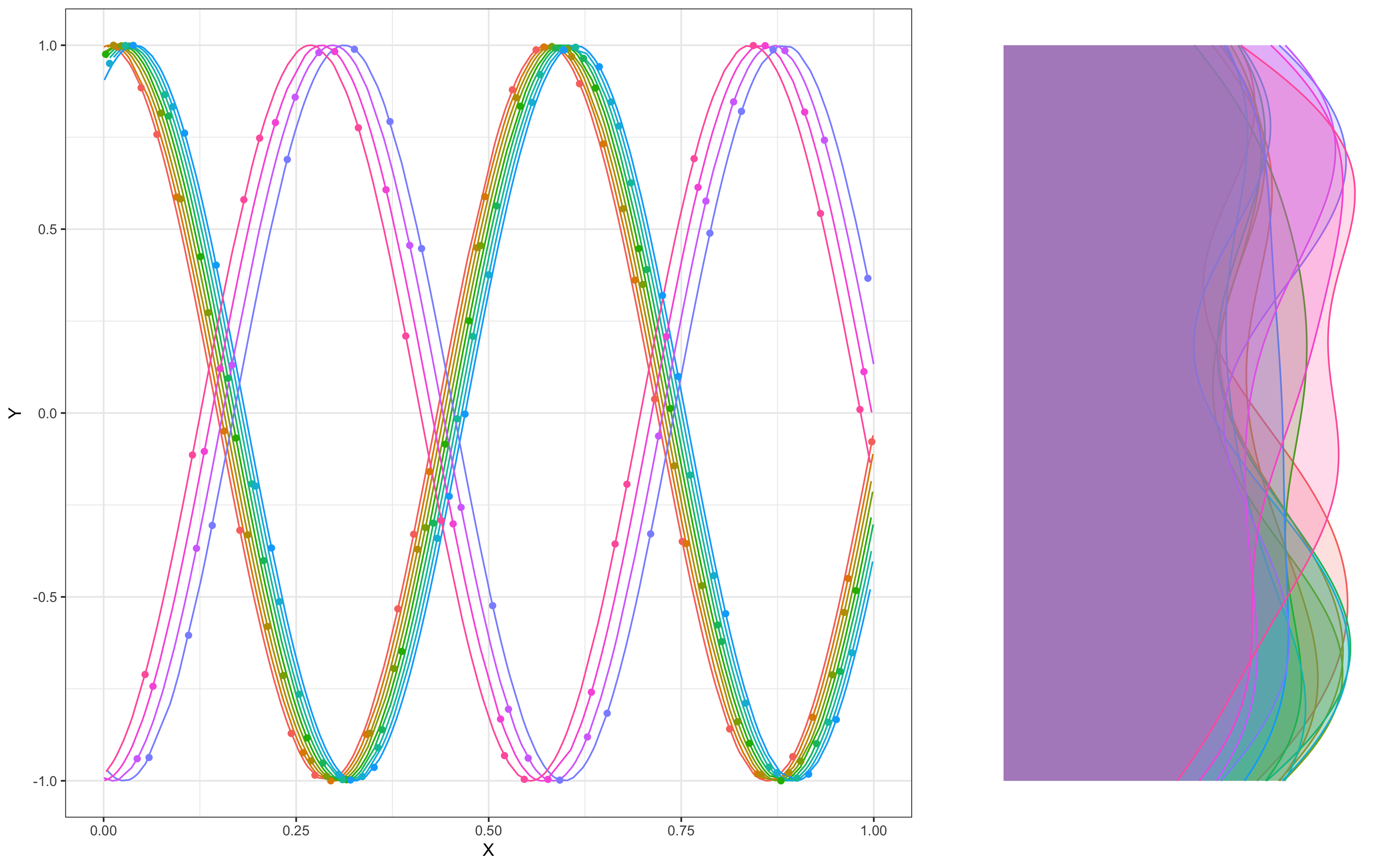} 
\caption{$f_1$ function: true function and training samples (left), output conditional probability for each level (right). Each level is represented by a different color.}
            \label{fig:f1_illustration}
\end{figure}

A potential limitation of target or distributional encoding may arise if the quantitative inputs do not have a strong main effect on the output. Indeed, we focus on characteristics of the empirical conditional distribution $\hat{P}^{Y}_{t,l}$ for each level $l$ of the input $u_t$: if there is little variation of this distribution when $l$ changes, the kernel may fail to capture relevant similarities between the levels. We illustrate in Figure \ref{fig:f1_illustration} such a setting on function $f_1$ from \citet{roustant2020} with one quantitative input and one qualitative input with $13$ levels.

Before discussing a workaround, we can first try to detect if a qualitative input has a small main effect. In computer experiments, this is achieved by computing \emph{first-order Sobol' indices} from global sensitivity analysis \citep{da2021basics}. Formally, the first-order Sobol' index of a random feature $Z$ on an output $Y$ is defined as
\begin{equation*}
S_Z = \frac{\textrm{Var}(\mathbb{E}(Y\vert Z))}{\textrm{Var}(Y)}
\end{equation*}
and quantifies the main effect of $Z$ on $Y$. As for estimation, for a quantitative input $X$, a practical estimator based on ranks was proposed by \citet{gamboa2022}:
\begin{eqnarray*}
\textrm{Var}(\mathbb{E}(Y\vert X)) &=& \mathbb{E}(\mathbb{E}(Y\vert X)^2) - \mathbb{E}(Y)^2\\
&\approx& \frac{1}{n} \sum_{i=1}^n y^{(i)} y^{(\sigma(i))} - \bar{y}_n^2
\end{eqnarray*}
with the permutation $\sigma(i)$ writes
\begin{align*}
\label{eq:perm}
\sigma(i) = \left\{
    \begin{array}{ll}
        \pi^{-1}(\pi(i)+1)  & \mbox{if } \pi(i)+1\leq n \\
        \pi^{-1}(1) & \mbox{otherwise}
    \end{array}
\right.
\end{align*}
where $\pi(i)$ is the rank of $x^{(i)}$ in the sample $(x^{(1)},\ldots,x^{(n)})$. For a quantitative input $U$, a straightforward Monte-Carlo estimator is given by
\begin{eqnarray*}
\textrm{Var}(\mathbb{E}(Y\vert U)) &=& \mathbb{E}(\mathbb{E}(Y\vert U)^2) - \mathbb{E}(Y)^2\\
&\approx& \sum_{l=1}^L \frac{N_l}{n} \bar{y}_{l}^2 - \bar{y}_n^2
\end{eqnarray*}
where we use the same notations as in Section \ref{sec:base_encod}. The former estimator can readily be used to detect if any quantitative input has a main effect, and as a consequence point towards a potential limitation of an encoding. Table \ref{tab:sobol_f1} gives the first-order sensitivity indices for function $f_1$: the main effect of the quantitative variable is clearly small, thus giving a hint that the interaction between features is strong.

\begin{table}[hbt!]
\centering
\begin{tabular}{|c|c|} 
 \hline
 $S_X$ & $S_U$ \\
  \hline
 $0.03$ & $0.01$  \\ 
 \hline
\end{tabular}
\caption{First-order indices for function $f_1$.}
\label{tab:sobol_f1}
\end{table}

Second-order Sobol' indices are precisely defined to measure such an interaction, and are defined as
\begin{equation*}
S_{Z_1,Z_2} = \frac{\textrm{Var}(\mathbb{E}(Y\vert Z_1,Z_2))-\textrm{Var}(\mathbb{E}(Y\vert Z_1))-\textrm{Var}(\mathbb{E}(Y\vert Z_2)) }{\textrm{Var}(Y)}.
\end{equation*}
Focusing on interactions between a quantitative and a qualitative input, we can extend the previous rank-based estimator: 
\begin{eqnarray*}
\textrm{Var}(\mathbb{E}(Y\vert X,U)) &=& \mathbb{E}(\mathbb{E}(Y\vert X,U)^2) - \mathbb{E}(Y)^2\\
&=& \mathbb{E}_U \left\{\mathbb{E}_X(\mathbb{E}(Y\vert X,U)^2)\right\}- \mathbb{E}(Y)^2\\
&\approx& \frac{1}{n}\sum_{l=1}^L  \sum_{i\in I_l} y^{(i)} y^{(\sigma_l(i))} - \bar{y}_n^2
\end{eqnarray*}
where $\sigma_l(i)$ now corresponds to the rank permutation of samples where $U=l$ only, denoted by $I_l$. In practice, the following strategy for handling interactions can then be employed:
\begin{enumerate}
    \item Compute first-order indices of all quantitative and qualitative inputs
    \item For all qualitative inputs:
    \begin{enumerate}
        \item If it has a sufficiently large main effect (e.g. $>5-10\%$), use the standard distributional encoding
        \item Otherwise, compute its second-order interaction with all quantitative inputs. For each significant interaction (e.g. $>5\%$), use distributional encoding on
        \begin{eqnarray*}
\hat{P}^{Y}_{l,s}=\frac{1}{N_{ls}} \sum_{i=1}^n \delta_{y^{(i)}}\mathds{1}_{u^{(i)}=l}\mathds{1}_{x^{(i)}\in I_s}
\end{eqnarray*}
where $\{I_s\}_{s=1}^S$ is a partition of size $S$ of the support of $X$ (obtained by e.g., empirical quantiles). 
    \end{enumerate}
    \item Assemble a product kernel with all these distributional encodings
\end{enumerate}
The advantage of this strategy is that interactions can be accounted for via step (b), but this requires a large training set and increases the problem dimension. We plan to investigate further its potential on large datasets in future work.

\newpage

\section{Additional numerical experiments}
\label{sec:addxp}

\begin{figure}[hbt!]

\begin{subfigure}{.475\linewidth}
  \includegraphics[width=\linewidth]{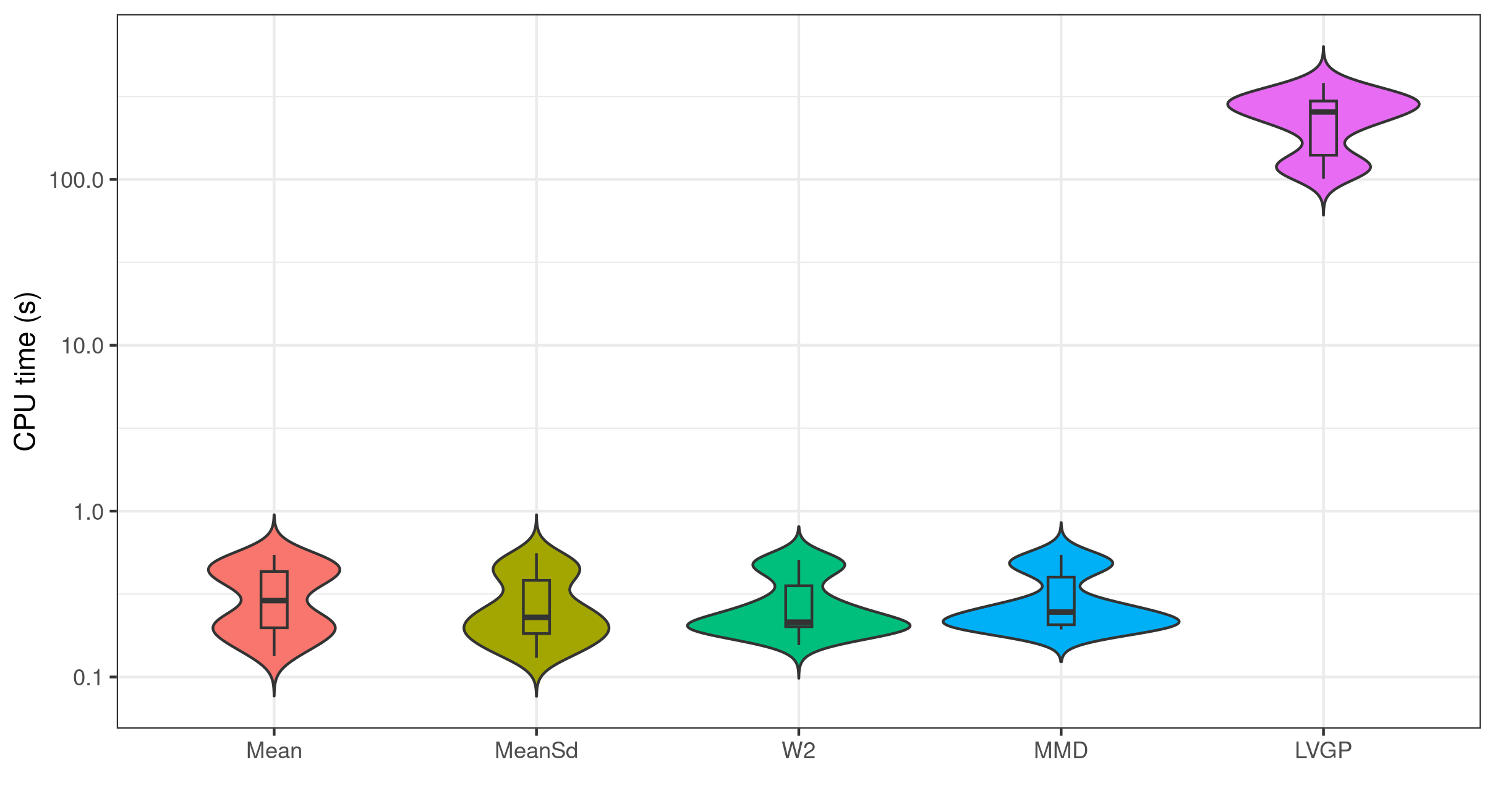}
  \caption{Beam bending.}
  \label{fig:bench_beam_time}
\end{subfigure}\hfill 
~ 
\begin{subfigure}{.475\linewidth}
  \includegraphics[width=\linewidth]{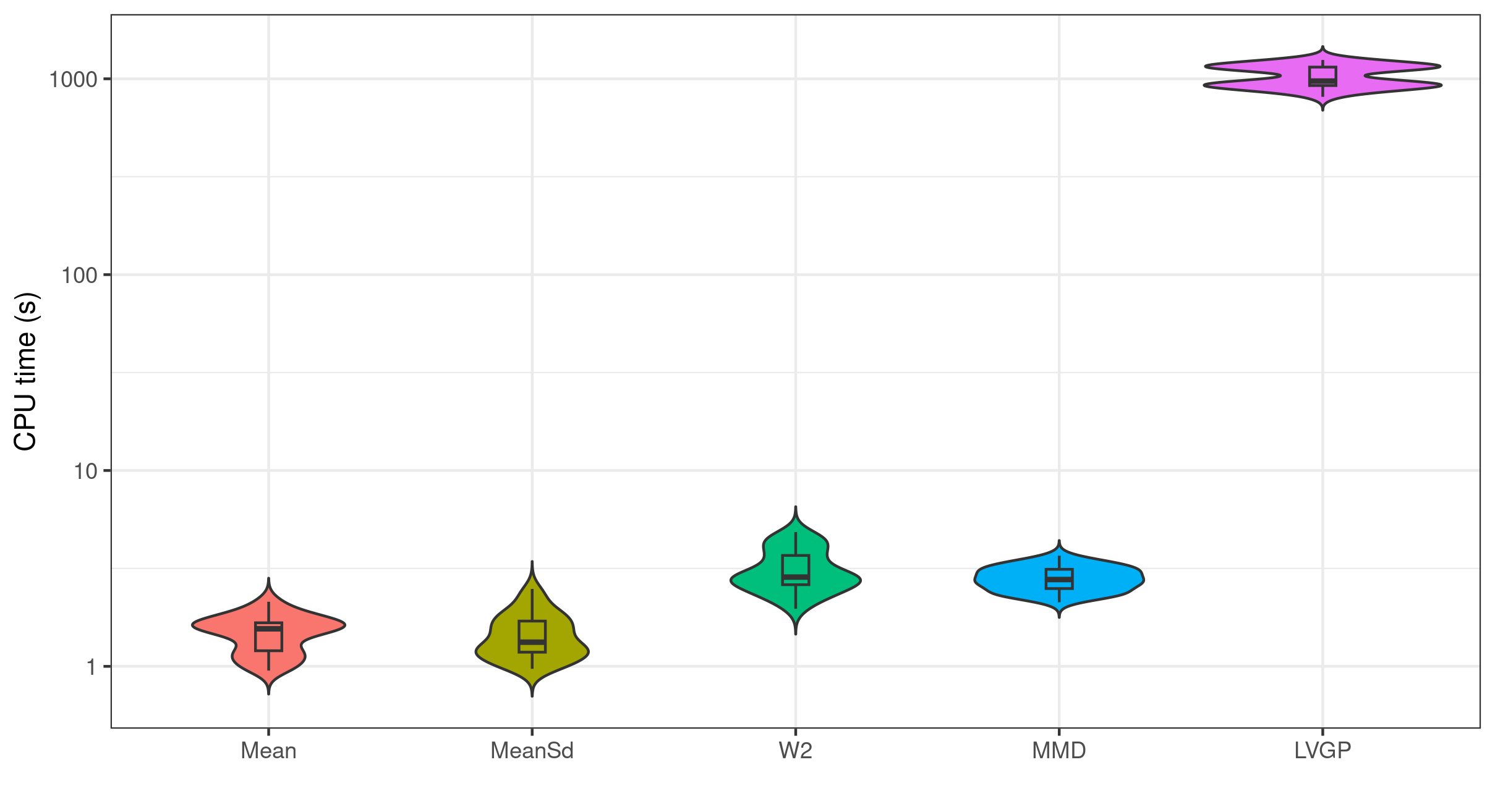}
  \caption{Borehole.}
  \label{fig:bench_borehole_time}
\end{subfigure}

\medskip 
\begin{subfigure}{.475\linewidth}
  \includegraphics[width=\linewidth]{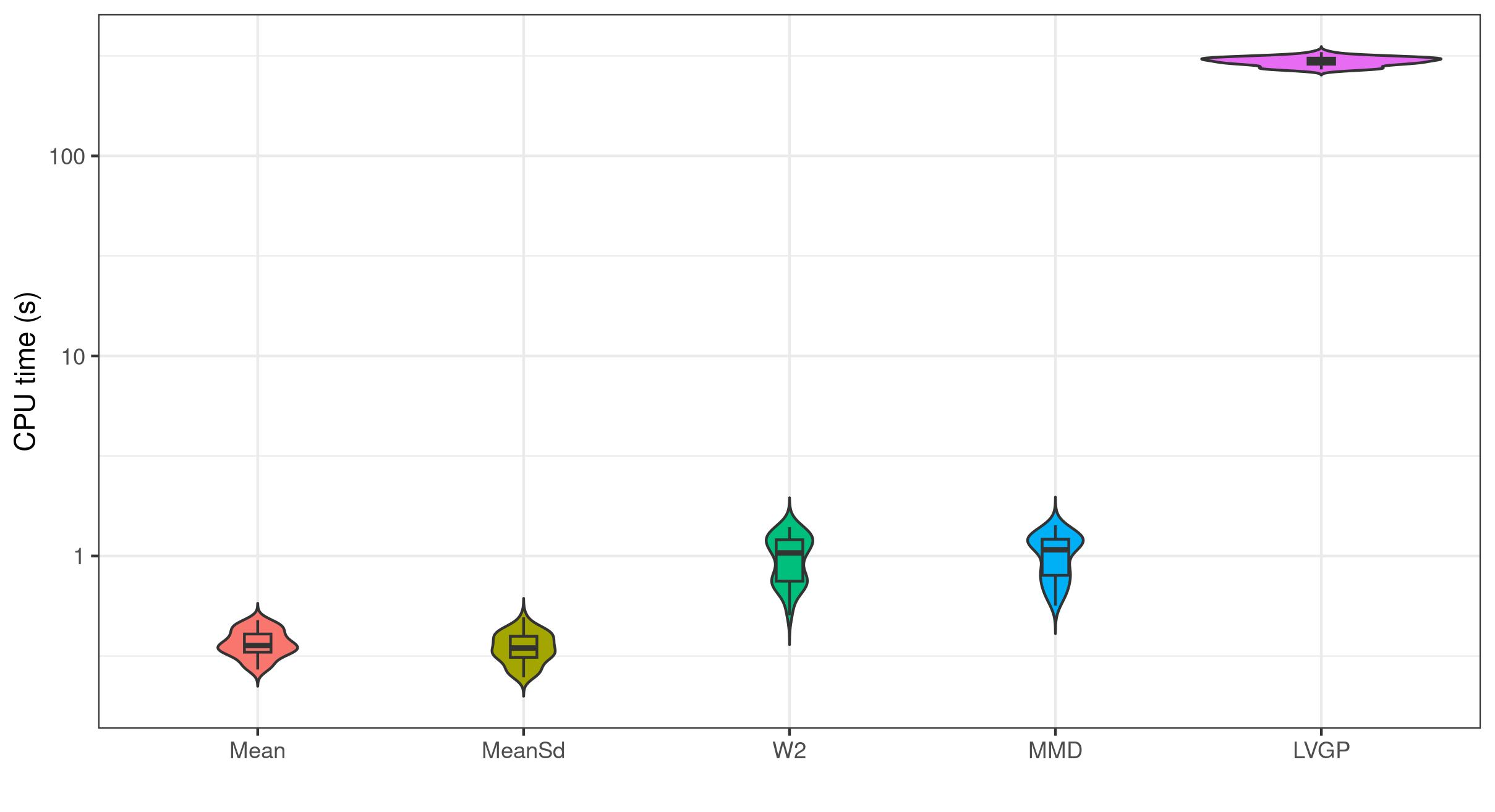}
  \caption{OTL.}
  \label{fig:bench_otl_time}
\end{subfigure}\hfill 
\begin{subfigure}{.475\linewidth}
  \includegraphics[width=\linewidth]{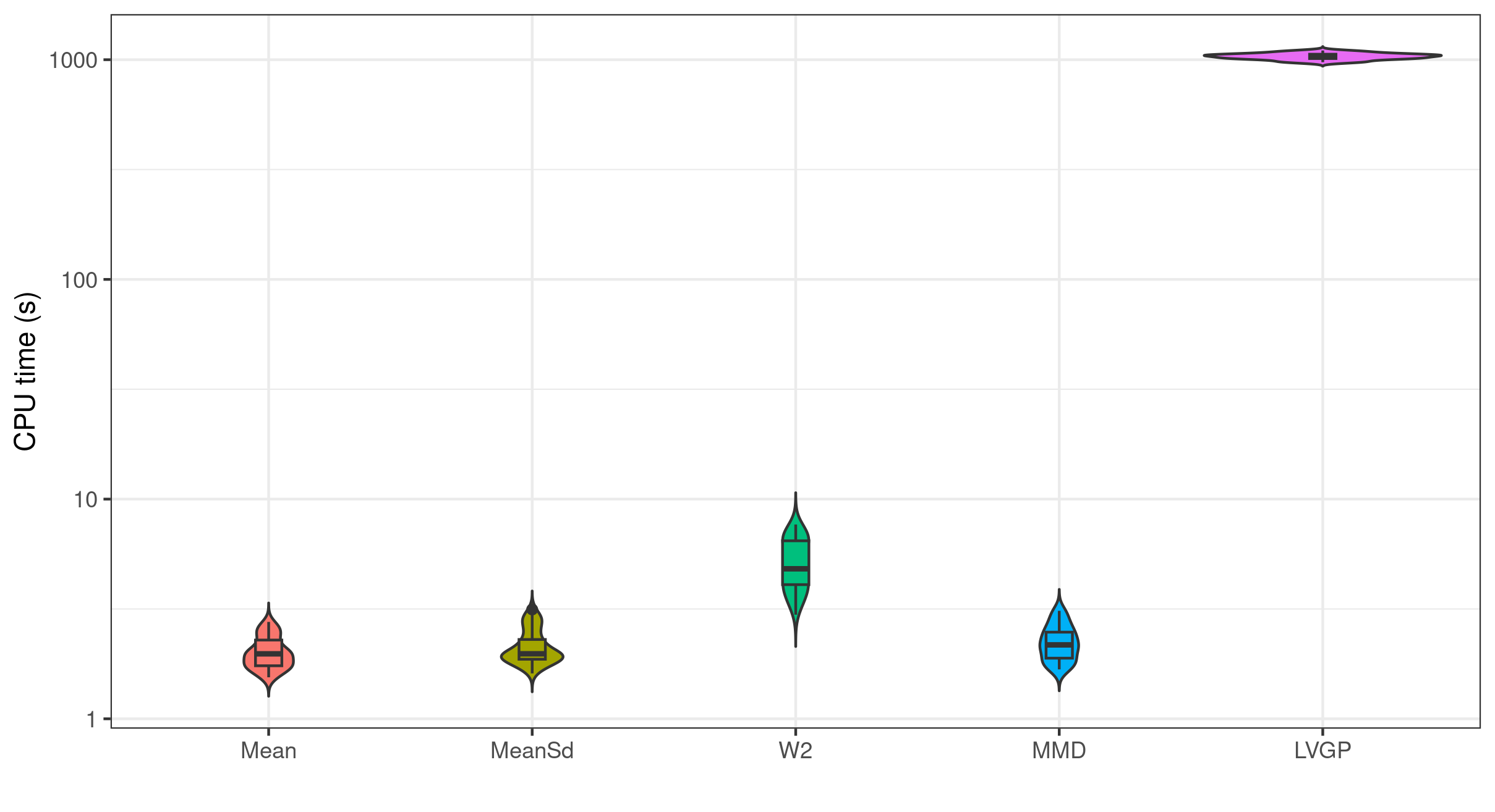}
  \caption{Piston.}
  \label{fig:bench_piston_time}
\end{subfigure}

\caption{Computation time on the four engineering problems with 50 replications.}
\label{fig:benchmark_time}
\end{figure}

\begin{figure}[hbt!]
\includegraphics[width=0.95\linewidth]{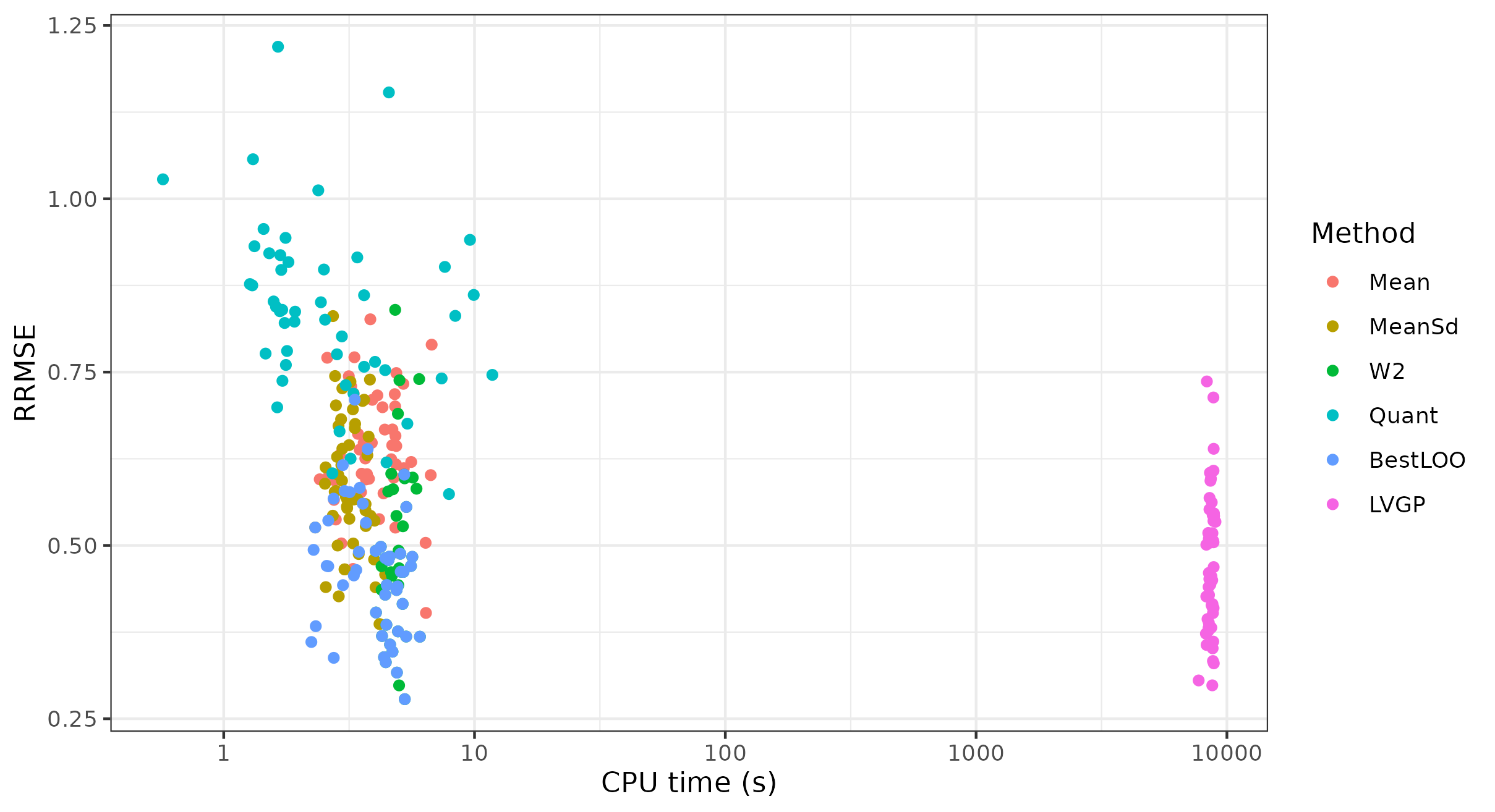}
\caption{RRMSE versus computation time on the material design problem with 50 replications.}
\label{fig:material_time}
\end{figure}

To complement the RRMSE results from Section \ref{sec:engxp}, we provide in Figures \ref{fig:benchmark_time} and \ref{fig:material_time} below the computation times for all methods with respect to the attained RRMSE. As already commented, distributional encoding and LVGP usually yield similar RRMSE, but the computational cost of LVGP is much higher.  

\begin{figure}[hbt!]
\includegraphics[width=0.95\linewidth]{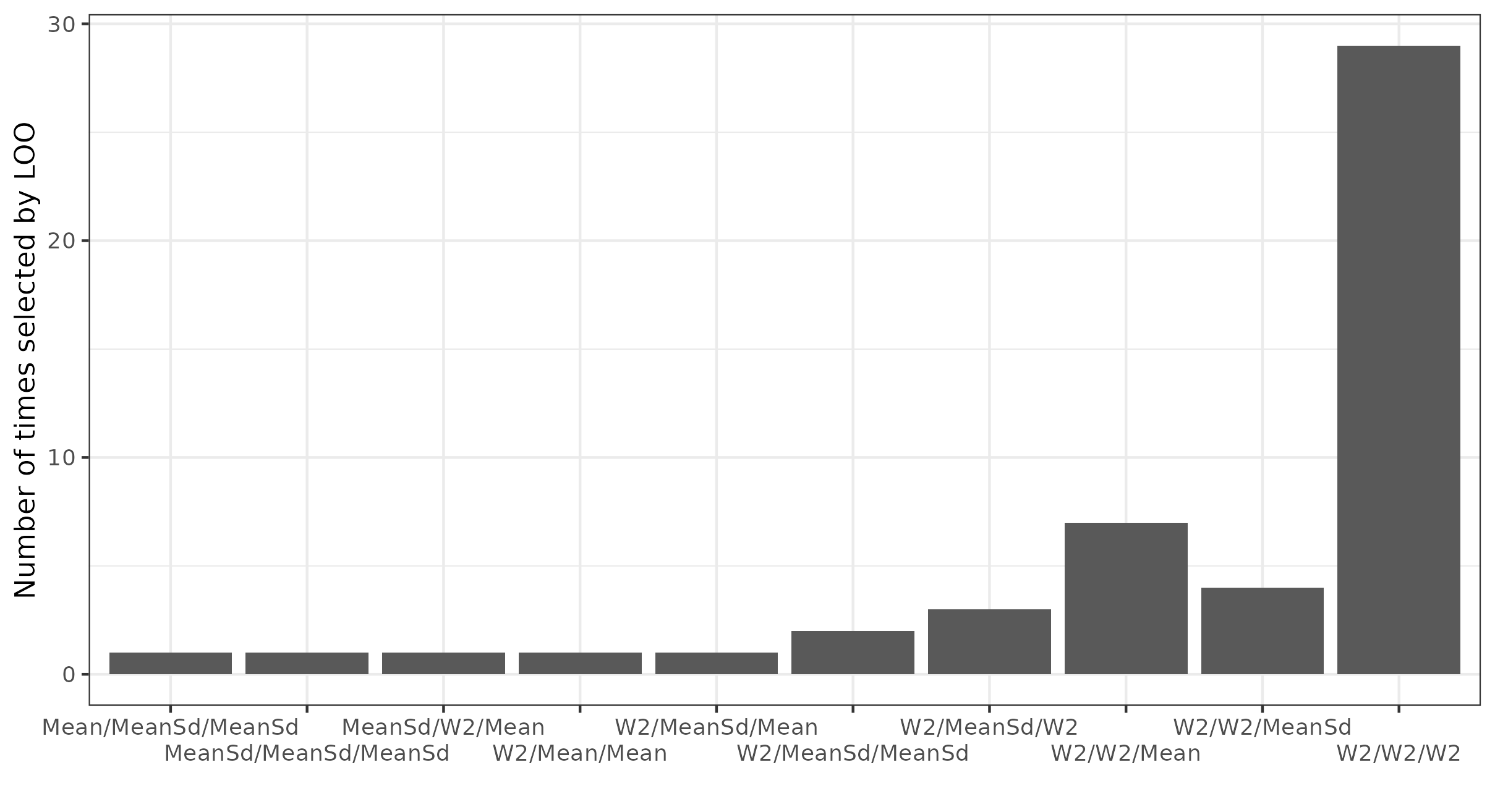}
\caption{RRMSE versus computation time on the material design problem with 50 replications.}
\label{fig:material_loo}
\end{figure}

For the material design case, we also report in Figure \ref{fig:material_loo} the results of the leave-one-out strategy, where for each replication we test all possible combinations of encodings for each qualitative input. Although the $W_2$ kernel is predominantly selected, we can observe that for the last qualitative feature, a simpler encoding is often sufficient.

 \end{document}